\documentclass[runningheads]{llncs}
\nonstopmode
 
\usepackage{eccv}

\usepackage{eccvabbrv}

\usepackage{amsfonts}
\usepackage{amsmath}
\usepackage{amssymb}
\usepackage{bbm}
\usepackage{tabularx}
\usepackage{tabularray}
\usepackage{enumitem}
\usepackage{graphicx}
\usepackage{xcolor}
\usepackage{pifont}
\usepackage{booktabs}
\usepackage{multirow}
\usepackage{wrapfig}

\usepackage[accsupp]{axessibility}  

\NewDocumentCommand{\todo}
{ mO{} }{\textcolor{magenta}{\textsuperscript{\textit{TODO}}\textsf{\textbf{\small[#1]}}}}

\usepackage{hyperref}

\usepackage{orcidlink}

\begin{document}

\title{Training-Free Model Merging for \\ Multi-target Domain Adaptation} 

\titlerunning{Training-Free Model Merging for Multi-target Domain Adaptation}


\author{Wenyi Li\textsuperscript{*} \inst{1} \and
Huan-ang Gao\textsuperscript{*} \inst{1} \and
Mingju Gao\inst{1} \and
Beiwen Tian\inst{1} \and
Rong Zhi\inst{2} \and
Hao Zhao\textsuperscript{\textdagger} \inst{1}}

\authorrunning{W. Li et al.}


\institute{Institute for AI Industry Research (AIR), Tsinghua University\and
Mercedes-Benz Group China Ltd. \\
\email{liwenyi19@mails.ucas.ac.cn, zhaohao@air.tsinghua.edu.cn}\\
\setcounter{footnote}{1}
\footnotetext{\textsuperscript{*} Indicates Equal Contribution. \textsuperscript{\textdagger} Indicates Corresponding Author.}
}

\maketitle

\begin{abstract}
In this paper, we study multi-target domain adaptation of scene understanding models.
While previous methods achieved commendable results through inter-domain consistency losses, they often assumed unrealistic simultaneous access to images from all target domains, overlooking constraints such as data transfer bandwidth limitations and data privacy concerns.
Given these challenges, we pose the question: How to merge models adapted independently on distinct domains while bypassing the need for direct access to training data?
Our solution to this problem involves two components, merging model parameters and merging model buffers (\text{\ie}, normalization layer statistics).
For merging model parameters, empirical analyses of mode connectivity surprisingly reveal that linear merging suffices when employing the same pretrained backbone weights for adapting separate models.
For merging model buffers, we model the real-world distribution with a Gaussian prior and estimate new statistics from the buffers of separately trained models.
Our method is simple yet effective, achieving comparable performance with data combination training baselines, while eliminating the need for accessing training data.
Project page: \url{https://air-discover.github.io/ModelMerging}.
\end{abstract}

\section{Introduction}

Scene understanding models need to perform reliably across various domains (\ie, diverse lighting \cite{zheng2023steps}, weather, and urban landscapes) to be truly useful for autonomous driving worldwide. 
The typical approach of supervised learning, however, relies heavily on costly pixel-level annotations by humans, which significantly hampers the scalability of these segmentation models. 
As such, the study of multi-target domain adaptation \cite{gholami2020unsupervised, nguyen2021unsupervised, isobe2021multi} (MTDA) is becoming increasingly significant. 
This area of research focuses on devising strategies to simultaneously utilize large-scale unlabeled real-world data from multiple domains along with labeled synthetic data during training \cite{wei2024editable}, providing a scalable approach to enhance the robustness of these models.

\begin{wrapfigure}[24]{r}{0.5\textwidth}
    \begin{minipage}[t]{0.5\textwidth}
        \includegraphics[width=\linewidth]{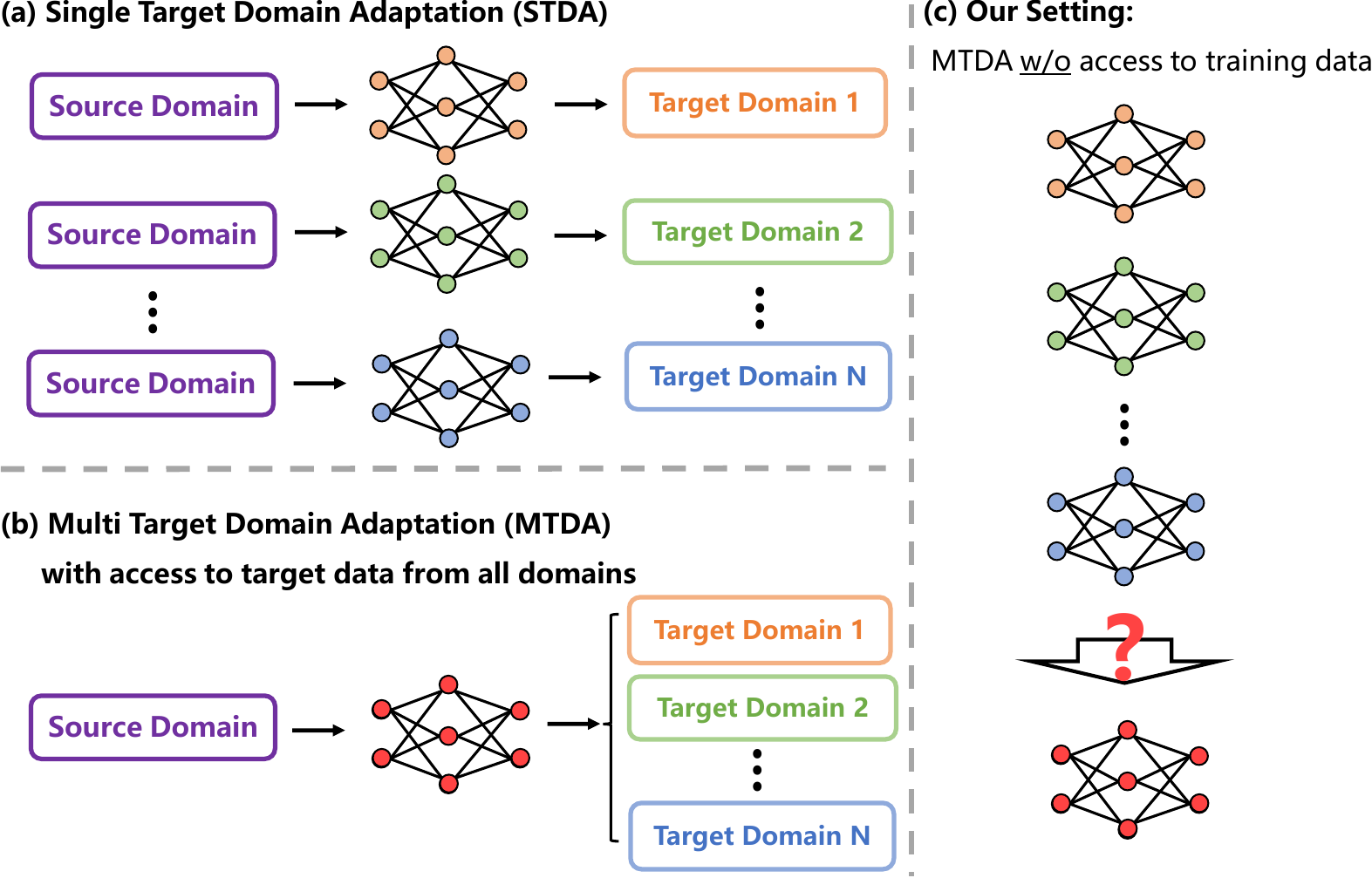}
        \caption{\textbf{Comparison of Domain Adaptation Settings.} (a) Single Target Domain Adaptation (STDA) focuses on leveraging labeled synthetic data and unlabeled data from a single target domain together for optimal performance in that target domain. (b) Multi-target Domain Adaptation (MTDA) with data access involves utilizing data from target domains together to train a single model capable of excelling across all these domains. (c) MTDA without direct access to training data, employing model merging to enhance robustness.}
        \label{fig:teaser}
    \end{minipage}
\end{wrapfigure}

MTDA presents a greater challenge compared to traditional single-target domain adaptation (STDA, illustrated in Fig.\ref{fig:teaser}(a)), due to the inherent difficulty in developing a single model that performs effectively across multiple target domains.
To tackle this, previous approaches \cite{gholami2020unsupervised, nguyen2021unsupervised, isobe2021multi, long2024adaptive} have employed consistency learning among various expert models and online knowledge distillation to construct a domain-generic student model. 
Nonetheless, a significant limitation of these methods is their reliance on simultaneous access to all target data, as illustrated in Fig.~\ref{fig:teaser}(b), which is usually impractical.
This impracticality stems partly from data transfer cost constraints, as a dataset comprising thousands of images can balloon to hundreds of gigabytes. Moreover, data privacy concerns are significant, with potential restrictions on the sharing or transferring of such data due to policies.

Facing these challenges, in this paper, we consider a novel problem setting illustrated in Fig.~\ref{fig:teaser}(c). 
Our primary focus remains on multi-target domain adaptation, but instead of accessing data from multiple domains, we gain access to \textit{models} that have been independently adapted to each of them. 
Our objective is to integrate these models into a single, effective model for various domains.

But how to merge multiple models into one while preserving their respective abilities on their respective domains? Our solution to this problem has two main components.
\textbf{(I) Merging model parameters. }
A straightforward method involves averaging the weights of models as a means of integration. 
Nevertheless, this approach raises questions about its underlying mechanisms and its reliability across different scenarios.
In our research, we conduct careful investigation into the conditions under which such merging is effective, and when it is not, by examining the concept of \textit{linear mode connectivity} of models. 
Through meticulous analysis, we find that pretraining significantly contributes to enhancing this linear connectivity among trained models.
\textbf{(II) Merging model buffers.} We identify the importance of model buffers, \ie, batch normalization (BN) layer statistics, which capture domain-specific characteristics in our multi-target domain adaptation setting. Leveraging the Gaussian prior assumption of BN layers, we estimate new means and variances for the merged layers based on the statistics of separately trained models.

Our method, simple yet effective, demonstrates notable improvements in performance compared to a variety of baselines.
For instance, when we apply our model merging technique to the state-of-the-art STDA method \cite{hoyer2022hrda} using a ResNet101 \cite{he2016deep} backbone, we observe a substantial increase of +5.6\% in harmonic mean of mIoUs for individual domains.
Remarkably, this level of performance rivals that achieved by baseline methods that involve training with multiple combined datasets, even when data availability is constrained. 
Furthermore, our technique outperforms previous top-performing multi-target domain adaptation methods that utilize explicit consistency training by a considerable margin, underscoring the critical role of exploring model connectivities.
In summary, our contributions are as follows:
\begin{itemize}
\item[$\bullet$] We conduct a systematic exploration of mode connectivity in domain-adapted scene parsing models, revealing the underlying conditions of when model merging works.
\item[$\bullet$] We introduce a model merging technique including parameter merging and buffer merging for multi-target domain adaptation tasks, applicable to \textit{any} single-target domain adaptation model.
\item[$\bullet$] Our approach achieves performance comparable to that of training with multiple combined datasets, even when data availability is constrained.
\end{itemize}

\section{Related Works}

\subsection{Domain Adaptation for Semantic Segmentation}

Unsupervised domain adaptation \cite{ganin2015unsupervised} helps to reduce annotation costs in pixel-wise semantic segmentation by utilizing unlabeled real-world images alongside labeled synthetic samples. 

\textbf{Single-target Domain Adaptation.}
The challenge of STDA mainly involves how to effectively reduce the distribution gap between labeled synthetic data and unlabeled single-domain real-world images. Methods to tackle this issue include domain-invariant representation learning \cite{yang2020label, li2019bidirectional, kim2020learning}, adversarial training \cite{gong2019dlow, hoffman2018cycada, ganin2016domain, hoffman2016fcns, long2018conditional, sajjadi2016regularization, tsai2018learning, vu2019advent},
and teacher-student self-training \cite{hoyer2022daformer, hoyer2022hrda, hoyer2023mic, gao2023semi, gao2023dqs3d, guan2023bridging}. 
Among these approaches, the last one has proven to be highly successful.
It leverages synthetic data \cite{tian2024latency, li2024fairdiff, gao2024scp, chen2024ultraman} as source domains, generates pseudo labels \cite{lee2013pseudo} for target domains and enforces consistency regularization over data augmentation strategies \cite{araslanov2021self, choi2019self, melas2021pixmatch, lai2021semi, zhang2021multiple, zheng2021rectifying, zhou2022uncertainty, hoyer2022daformer, hoyer2023improving, tian2023unsupervised}.
Despite STDA being a broadly studied topic, it has limitations for real-world applications like autonomous driving, where vehicles encounter diverse road scenarios beyond a single domain.

\textbf{Multi-target Domain Adaptation.}
MTDA presents greater challenges than STDA, particularly in terms of integrating domain-specific knowledge from multiple sources into a single model. 
While initial efforts in this area have focused on classification tasks \cite{gholami2020unsupervised, nguyen2021unsupervised, yu2018multi}, MTDA has also been explored for semantic segmentation \cite{nguyen2021unsupervised, isobe2021multi}, employing strategies like inter-domain consistency regularization and online knowledge distillation. 
Nonetheless, the application of these explicit inter-domain regularization techniques limits the use of certain STDA strategies, like multi-resolution training, due to the excessive GPU memory demands associated with managing multiple student models. Consequently, these approaches \cite{nguyen2021unsupervised, isobe2021multi} significantly underperform when compared to STDA methodologies \cite{hoyer2022daformer, hoyer2022hrda, hoyer2023mic}. 
Moreover, they rely on the impractical assumption of having simultaneous access to data from all target domains.
So we ask the question: is it possible to train several models under the best STDA setting and merge them without access to training data (\ie, in a zero-shot way \cite{he2024zero})?

\subsection{Multi-target Learning with Constrained Data Assumption}
\textbf{Federated Learning (FL). }
While FL\cite{mcmahan2017communication} focuses on the \emph{joint training} of models, our research studies the direct merging of models \emph{post training}.
Representative FL methods confront challenges including communication cost \cite{mishchenko2022proxskip}, data heterogeneity\cite{li2020federated}, and support for device capabilities\cite{horvath2021fjord} in a distributed setting. In contrast, our approach diverges in methodology by leveraging the advancements in large-scale, centralized pre-training of vision models. We answer the question that \textit{under which specific conditions} can we simply \textit{merge} pre-trained models adapted to distinct data domains.
\textbf{Federated MTDA.} In this setting \cite{yao2022federated}, the distributed client data is unlabeled, and a centralized labeled dataset is available on the server. In our STDA stage, labeled dataset is accessible for each domain (comparable to clients). 
\textbf{Class-Incremental Domain Adaptation (CIDA).} CIDA \cite{kundu2020class} learns novel target-domain classes in a domain shift-aware manner, while they do not touch \textit{multiple target} domains.

\subsection{Mode Connectivity for Neural Networks}

\textbf{Mode connectivity} refers to the fact where local minima in the loss functions of neural networks can be connected through a curve in the parameter space, along which the performance does not experience significant deterioration \cite{garipov2018loss, tatro2020optimizing}.
Linear mode connectivity is a stronger constraint requiring that the convex combination of two minima stay in the same loss basin, which is closely related to the lottery ticket hypothesis \cite{frankle2018lottery} and has direct implications for continual learning \cite{mirzadeh2020linear}.
Exploiting the property of linear mode connectivity, several works \cite{qin2022exploring, neyshabur2020being, huang2017snapshot, izmailov2018averaging, von2020neural, wortsman2022robust} integrate weights of neural networks for a more robust merged model.
In our study, we also observe a similar phenomenon in models adapted to different domains, and we conduct an empirical analysis to trace the origin of it. 
We also identify the importance of batch normalization layer statistics in the multi-target domain adaptation context, as the diverse characteristics inherent in multiple domains are distinctly captured in these statistical layers.
Different from the aforementioned work focused on language or image inputs from subsets of the same dataset, we propose a novel method to effectively merge these statistics.

\section{Methodology}

\subsection{Overview}

\begin{wrapfigure}[21]{R}{0.55\textwidth}
    \begin{minipage}[t]{0.55\textwidth}
    \centering
    \includegraphics[width=1.0\linewidth]{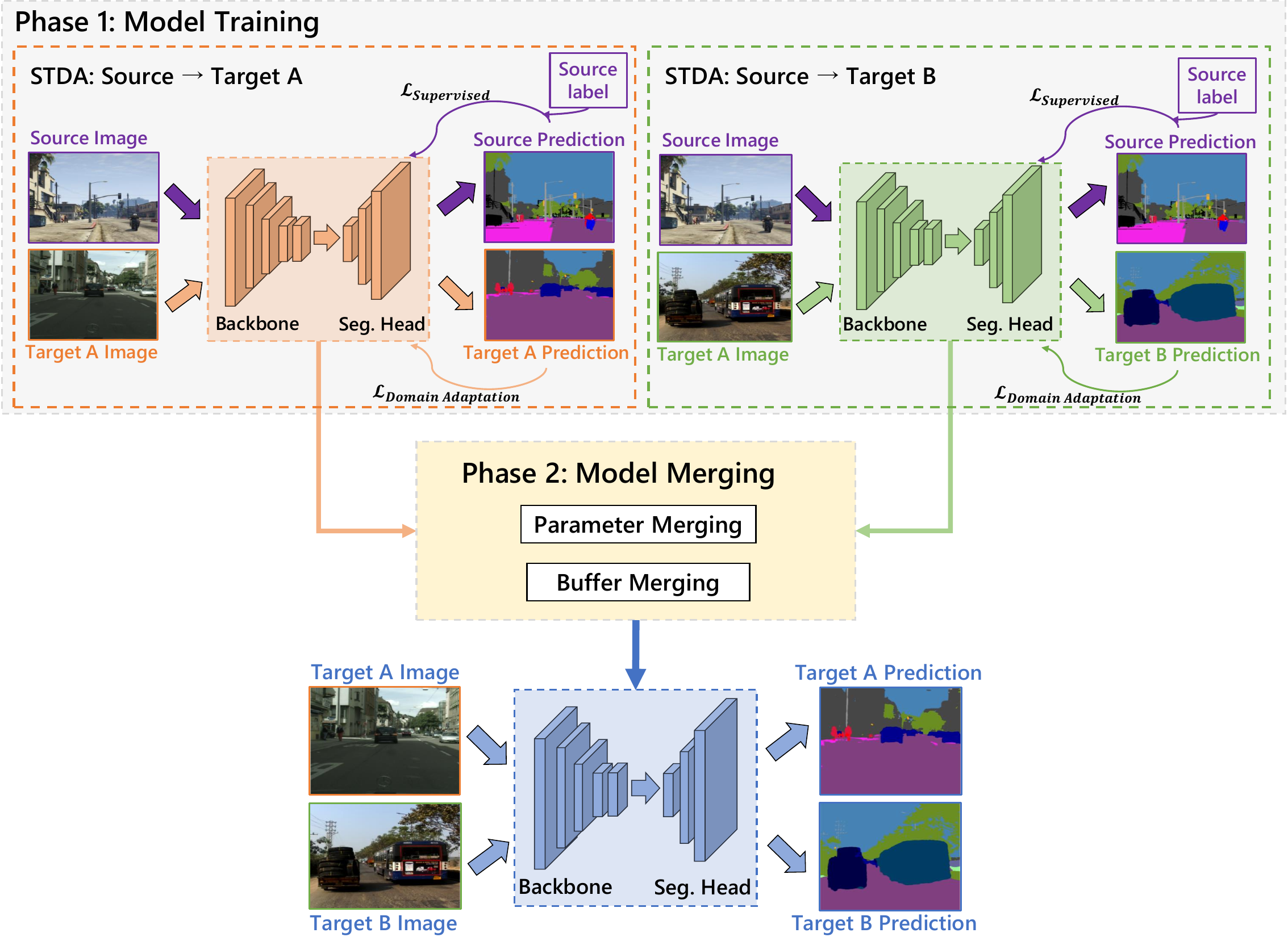}
    \caption{\textbf{Overview of Two-stage Pipeline of Our Proposed Multi-target Domain Adaptation Solution.} After training STDA methods on separate domains, we integrate models together using our proposed merging techniques. }
    \label{fig:pipeline}
    \end{minipage}
\end{wrapfigure}

We introduce a novel pipeline designed to address the challenging problem of multi-target domain adaptation in semantic segmentation, which is illustrated in Fig.~\ref{fig:pipeline}. What sets our approach apart from previous methods \cite{gholami2020unsupervised, nguyen2021unsupervised, isobe2021multi, ashmore2015method, yurochkin2019bayesian, wang2020federated} is the elimination of the impractical assumption that images of all target domains concurrently accessible during the adaptation phase. Instead, our pipeline comprises two distinct phases: a single-target domain adaptation phase and a model merging phase.
In the first phase, we train models adapted for individual target domains separately, while the second phase focuses on merging these adapted \textit{models} together to create a robust model, \textbf{without access to any training data.}

Our primary focus lies on the proposed model merging phase. In the initial phase, we simply adopt the state-of-the-art unsupervised domain adaptation approach, HRDA \cite{hoyer2022hrda}, leveraging various backbone architectures such as ResNet \cite{he2016deep} and vision transformer \cite{xie2021segformer}.
Our approach encompasses two critical components of the models: parameters (\ie, weights and biases for the learnable layers) and buffers (\ie, running statistics for the normalization layers).
We provide detailed explanations of the merging techniques and the underlying motivations for these techniques in Sec.~\ref{param-method} and Sec.~\ref{buffer-method} for merging parameters and buffers, respectively.

\subsection{Merging Parameters}
\label{param-method}

\begin{figure}[!t]
	\centering
	\includegraphics[width=0.7\linewidth]{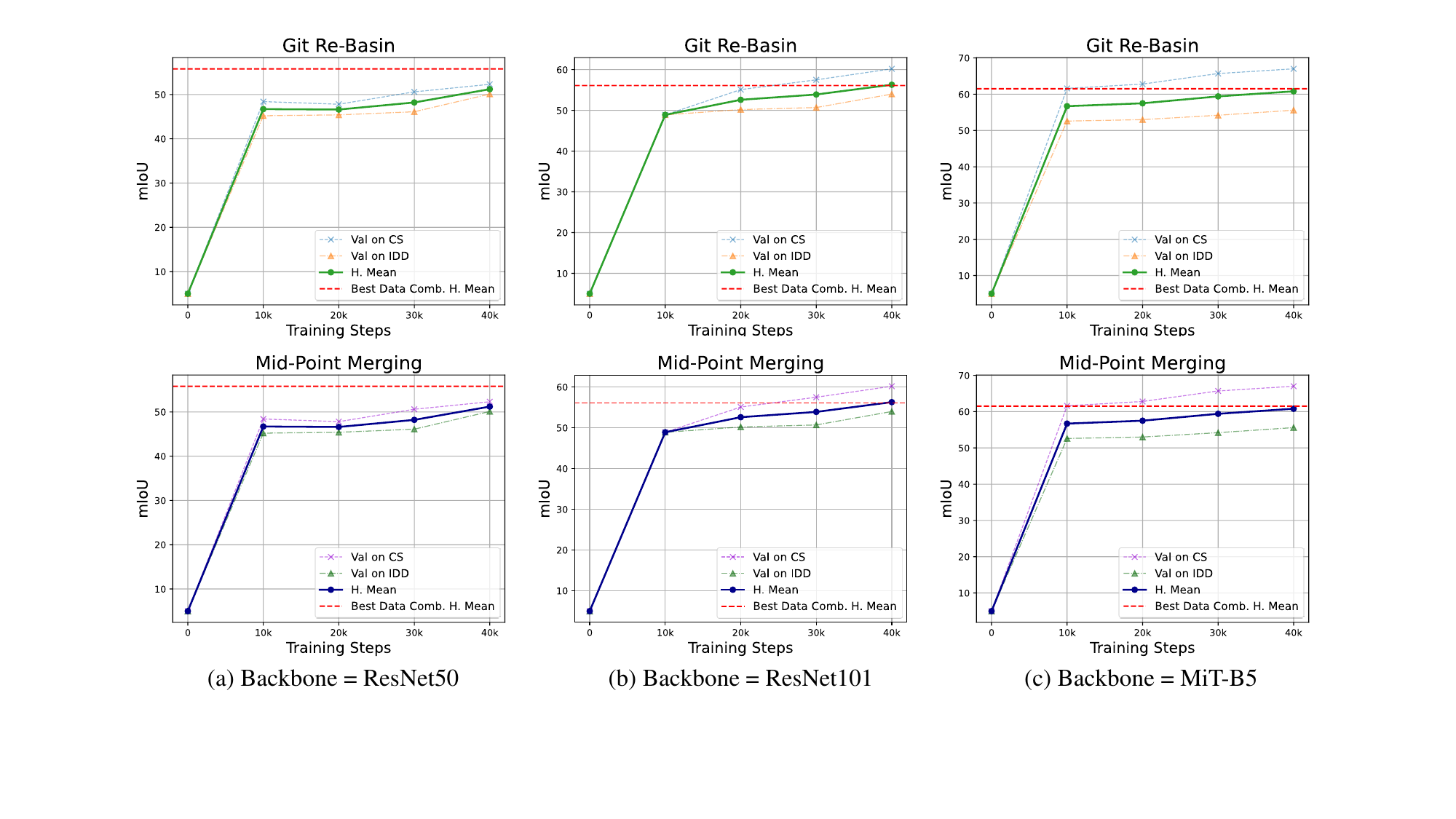}
	\caption{\textbf{Results of Git Re-Basin and Mid-Point Merging on Different Backbones.} In our domain adaptation scenario, Git Re-Basin \cite{ainsworth2022git} reduced to a straightforward mid-point merging approach.}
    \label{fig:midpoint_rebasin}
\end{figure}

\begin{figure}[!t]
	\centering
	\includegraphics[width=0.7\linewidth]{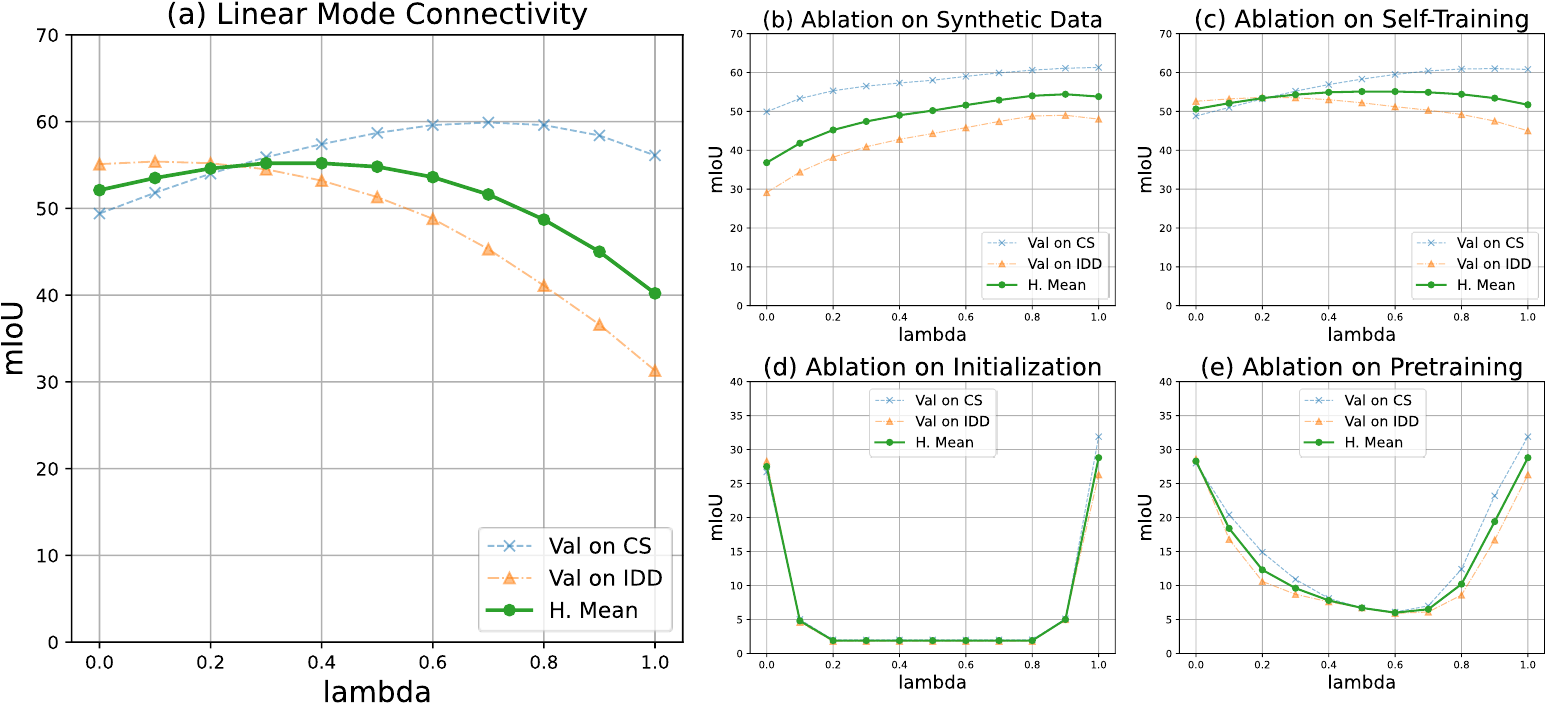}
    \caption{\textbf{Empirical Analysis for Linear Mode Connectivity.} (a) Exploring the linear mode connectivity of two trained ResNet101 backbones targeted at two different domains. (b-e) Ablation studies on synthetic data, self-training architecture, initializaiton weights and pretrained weights to find the cause of the linear mode connectivity.}
    \label{fig:linear-analysis}
\end{figure}

\textbf{Permutation-based methods degenerate.}
In fact, the idea of merging the weights and biases of learnable layers between models has been a frontier research area. 
Among these efforts, a particularly promising line of research has emerged, known as permutation-based methods \cite{entezari2021role, ainsworth2022git, jordan2022repair, pena2023re}. These methods operate under the assumption that, when accounting for all potential permutation symmetries of units in hidden layers of neural networks, the loss landscapes typically form a single basin. 
Therefore, when merging model parameters $\mathbf{\Theta}_A$ and $\mathbf{\Theta}_B$, the primary goal of these methods is to find a set of permutation transformations $\pi(\cdot)$ that ensures $\pi(\mathbf{\Theta}_B)$ is functionally equivalent to $\mathbf{\Theta}_B$, while also residing within an approximately convex basin near the reference model $\mathbf{\Theta}_A$.
After that, with a simple mid-point merging ($\lambda = \frac{1}{2}$ in Eq.~\ref{linear-merge}), we can acquire a merged model $\mathbf{\Theta}'$ that exhibits better generalization ability than single models,
\begin{equation}
    \label{linear-merge}
    \mathbf{\Theta}' = \lambda \mathbf{\Theta}_A + ( 1 - \lambda ) \pi(\mathbf{\Theta}_B).
\end{equation}

In our scenario, both $\mathbf{\Theta}_A$ and $\mathbf{\Theta}_B$ are trained during the first phase employing identical network architectures \cite{hoyer2022hrda} and utilizing the same synthetic images and labels \cite{richter2016playing}. However, they are adapted to samples from distinct domains for semantic segmentation \cite{cordts2016cityscapes, varma2019idd}.
Our initial attempt involved employing a representative permutation-based method known as Git Re-Basin \cite{ainsworth2022git}. This method transforms the task of finding permutation-symmetric transformations into a linear assignment problem (LAP), for which efficient and practical algorithms are available. Surprisingly, in our experimental setup, Git Re-Basin's performance equaled that of a simple mid-point merging across all network architectures, including ResNet50 \cite{he2016deep}, ResNet101 \cite{he2016deep}, and MiT-B5 \cite{xie2021segformer}, and the results of mid-point merging (also results of Git Re-Basin \cite{ainsworth2022git}) are illustrated in Fig.~\ref{fig:midpoint_rebasin}.
Further investigation revealed that the permutation transformation discovered by Git Re-Basin \cite{ainsworth2022git} remained identical permutation through iterations of solving LAP, suggesting that in our domain adaptation scenario, Git Re-Basin \cite{ainsworth2022git} reduced to a straightforward mid-point merging approach.

\textbf{Empirical analysis of linear mode connectivity.}
We further investigate the above degeneration problem through the lens of \textit{linear mode connectivity} \cite{garipov2018loss, frankle2020linear, mirzadeh2020linear}.
Specifically, we use continuous curve $\phi(\lambda) : [0, 1] \rightarrow \mathbb{R}^{|\mathbf{\Theta}|}$ to connect $\mathbf{\Theta}_A$ and $\mathbf{\Theta}_B$ in the parameter space. In this specific case, we consider a linear path as follows,
\begin{equation}
    \phi(\lambda) = \lambda \mathbf{\Theta}_A + ( 1 - \lambda ) \mathbf{\Theta}_B.
\end{equation}

After defining the curve connecting the models to be merged, we traverse along the curve and evaluate the performance of the interpolated models. To gauge the effectiveness of these models in adapting to the two specified target domains, denoted as $\mathcal{D}_1$ and $\mathcal{D}_2$, respectively, we employ the $\text{Harmonic Mean}$ as our primary evaluation metric,
\begin{equation}
    \text{Harmonic Mean} = \frac{2 \cdot \text{mIoU}_{\mathcal{D}_1} \cdot \text{mIoU}_{\mathcal{D}_2}}{\text{mIoU}_{\mathcal{D}_1} + \text{mIoU}_{\mathcal{D}_2}}.
\end{equation}

We select harmonic mean as the metric as it gives more weight to smaller values, which corresponds to the worst-case performance among various cities over the world. It effectively penalizes scenarios where performance in one domain (e.g., in well-developed big cities) is disproportionately high while other domains (e.g., in a rural third-world town) have low performance. We believe this is aligned with the initial goal of the multi-target domain adaptation task. On the contrary, in an extreme case, scoring 100\% in one city and zero on all other three cities still leads to 25\% performance using arithmetic mean numbers, which we believe is of limited value in the multi-target domain adaptation problem.

The evaluation results of the interpolation are depicted in Fig.~\ref{fig:linear-analysis}(a). 'CS' and 'IDD' denotes target datasets Cityscapes \cite{cordts2016cityscapes} and Indian Driving Dataset \cite{varma2019idd}, respectively.
Notably, it is evident that the two models from the first phase are already linearly mode connected without permutation, as the harmonic mean of the interpolated models outperforms the performance of individual models for both domains.

\textbf{Understanding the cause of linear mode connectivity.}
Given the aforementioned revelation, we inquire: What is the underlying reason behind the linear mode connectivity property observed in previous domain adaptation methods? Subsequently, we conduct ablation experiments to investigate several constant factors during the training of $\Theta_A$ and $\Theta_B$ in the first phase.

\textit{Synthetic Data.}
The utilization of the same synthetic data may serve as a bridge between the two domains. 
To assess this, we partition the training data from synthetic set \cite{richter2016playing} into two distinct non-overlapping subgroups, each comprising 30\% of the original training samples. 
During the partitioning process, we group images with identical scene identifiers provided by the synthetic dataset into the same subgroup, while scenes with significant differences are placed in separate subgroups.
We train two single-target domain adaptation models, using the source domain provided by these two distinct subgroups, while setting the target domain as the CityScapes \cite{cordts2016cityscapes} dataset. We subsequently investigate the linear mode connectivity of the two resulting models.
The results, displayed in Fig.~\ref{fig:linear-analysis}(b), reveal that there is no notable decline in performance along the linear curve connecting the two resulting models within the parameter space. This observation suggests that the usage of the same synthetic data is not the primary factor influencing the linear mode connectivity.

\textit{Self-training Architecture.}
The utilization of a teacher-student self-teaching architecture \cite{tarvainen2017mean} may confine resulting models to the same basin within the loss landscape.
To assess this possibility, we disable the exponential moving average (EMA) update for the teacher models. Instead, we copy the student weights to teacher models during each iteration.
Subsequently, we proceed to train two single-target domain adaptation models, utilizing GTA \cite{richter2016playing} as the source domain and Cityscapes \cite{cordts2016cityscapes} and IDD \cite{varma2019idd} as the target domains, respectively. We then investigate the linear curve that connects the two resulting models within the parameter space, and the outcomes are presented in Fig.~\ref{fig:linear-analysis}(c).
We can see that the linear mode connectivity property remains intact.

\textit{Initialization and Pretraining.}
The practice of initializing the backbone with the same pretrained weights can potentially place it in a basin that is challenging to escape from during the training process. To examine this potential scenario, we initialize two separate backbones with distinct weights and then proceed with domain adaptation targeting the Cityscapes \cite{cordts2016cityscapes} and IDD \cite{varma2019idd} domains.
During the evaluation of the linear interpolated models between the two converged models, we observe a notable deterioration in performance, as shown in Fig.~\ref{fig:linear-analysis}(d). 
To gain a deeper understanding of the underlying factors, we explore whether it is the identical initial weight or the pretraining process that contributes to this effect.
We initiate two backbones with the same weights but without pretraining and conduct the experiment once more. 
Interestingly, we still encounter a substantial performance barrier along the linear connecting curve in the parameter space, as reported in Fig.~\ref{fig:linear-analysis}(e). 
This implies that it is the pretraining process that plays a pivotal role in facilitating linear mode connectivity in the fine-tuned models.

\textbf{Summary.}
Our empirical analysis highlight that when commencing from identical pretrained weights, domain adaptation models can effectively transition to diverse target domains while still maintaining linear mode connectivity in the parameter space. 
Consequently, a straightforward mid-point merge between these trained models could generate models with robustness in both domains.

\subsection{Merging Buffers}
\label{buffer-method}

\begin{wrapfigure}[17]{R}{0.3\textwidth}
    \includegraphics[width=\linewidth]{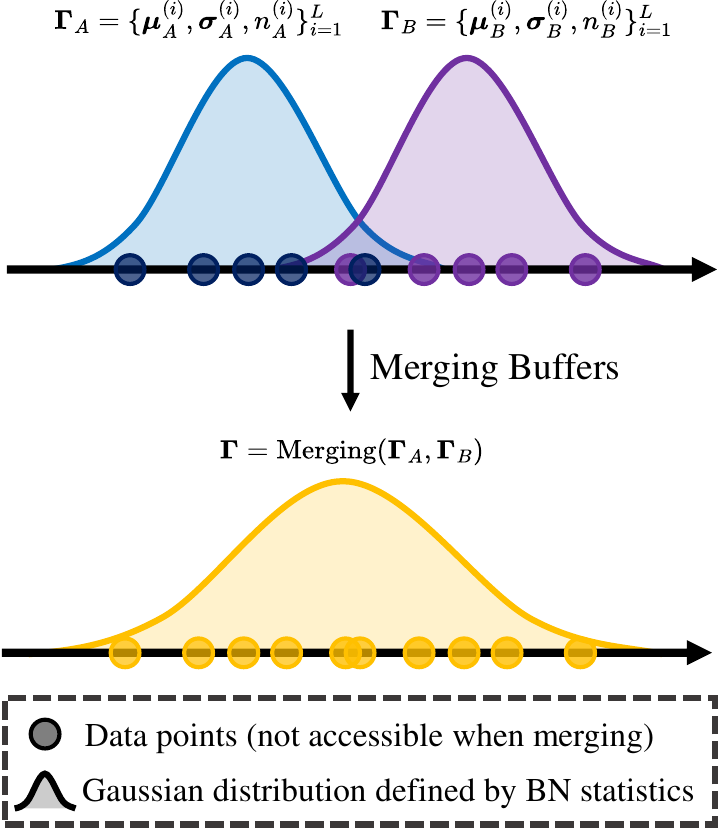}
    \caption{Illustration on Merging Statistics in Batch Normalization (BN) Layers.}
    \label{fig:buffer-merging}
\end{wrapfigure}

Buffers, namely running mean and variance for batch normalization (BN) layers, bear a close association with domains as they encapsulate domain-specific characteristics.
The question of how to effectively merge buffers when merging models is usually overlooked, as existing methods \cite{huang2017snapshot, izmailov2018averaging, von2020neural, wortsman2022robust} predominantly address the merging of two models trained on separate subsets within the same domain.
These approaches are reasonable, as buffers from any given model can be regarded as an unbiased estimation of the entire population, albeit solely derived from randomized data subsamples.
Nevertheless, in our specific problem context, we are investigating the merging of two models trained in completely distinct target domains, rendering the buffer merging problem non-trivial anymore.

Since we assume no access to any form of training data during the merging phase of model A and model B, our available information is confined to the set of buffers $\mathbf{\Gamma} = \{\boldsymbol{\mu}^{(i)}, \boldsymbol{\sigma}^{(i)}, n^{(i)}\}_{i=1}^{L}$. Here, $L$ represents the number of BN layers, while $\boldsymbol{\mu}^{(i)}$, $\boldsymbol{\sigma}^{(i)}$, and $n^{(i)}$ respectively signify the running statistics for mean, standard deviation, and the number of tracked batches for the $i$-th layer.
The statistics of the resulting BN layer is given as,
\begin{center}
\begin{equation}
\label{buffer-merging}
\resizebox{.5\linewidth}{!}{
\begin{math}
\begin{split}
        n^{(i)} &= n^{(i)}_A + n^{(i)}_B, \\
        \boldsymbol{\mu}^{(i)} &= \frac{1}{n^{(i)}} (n^{(i)}_A \boldsymbol{\mu}^{(i)}_A + n^{(i)}_B \boldsymbol{\mu}^{(i)}_B), \\
        [\boldsymbol{\sigma}^{(i)}]^2 &= \frac{1}{n^{(i)}}\Bigg(
        n^{(i)}_A [\boldsymbol{\sigma}^{(i)}_A]^2 + 
        n^{(i)}_B [\boldsymbol{\sigma}^{(i)}_B]^2 \\ &+ n^{(i)}_A [\boldsymbol{\mu}^{(i)} - \boldsymbol{\mu}^{(i)}_A]^2 +
        n^{(i)}_B [\boldsymbol{\mu}^{(i)} - \boldsymbol{\mu}^{(i)}_B]^2
        \Bigg).
\end{split}
\end{math}
}
\end{equation}
\end{center}

The rationale behind Eq.~\ref{buffer-merging} can be elucidated as follows.
BN layers are introduced to alleviate the issue of \textit{internal covariate shift} \cite{ioffe2015batch, santurkar2018does, bjorck2018understanding}, where the means and variances of inputs undergo changes as they pass through internal learnable layers.
In this context, our fundamental consideration is that subsequent learnable layers anticipate the output of the merged BN layer to follow a normal distribution. 
Since the resulting BN layer hold the the inductive bias of inputs conforming to a Gaussian prior, we estimate $\boldsymbol{\mu}^{(i)}$ and $[\boldsymbol{\sigma}^{(i)}]^2$ from what we get from $\mathbf{\Gamma}_A$ and $\mathbf{\Gamma}_B$.
As depicted in Figure~\ref{fig:buffer-merging}, we are provided with two sets of means and variances of data points sampled from this Gaussian prior, along with the sizes of these sets. We leverage these values collectively to estimate the parameters of this distribution.

\textbf{Extending to more domains.} When extending the merging method to $m (m \geq 2)$ Gaussian distributions, the number of tracked batches $ n^{(i)}$, the weighted average of the means $\boldsymbol{\mu}^{(i)}$ and the weighted average of the variances can be calculated as follows.

\begin{equation}
\label{m-buffer-merging-n-and-mean}
\begin{split}
        n^{(i)} =& n^{(i)}_1 + n^{(i)}_2 + \cdots +n^{(i)}_M, \\
        \boldsymbol{\mu}^{(i)} =& \frac{1}{n^{(i)}} (n^{(i)}_1 \boldsymbol{\mu}^{(i)}_1 + n^{(i)}_2 \boldsymbol{\mu}^{(i)}_2 + \cdots + n^{(i)}_M \boldsymbol{\mu}^{(i)}_M),\\
        \boldsymbol{\sigma}^2 =& \frac{\sum_{j=1}^{M} n^{(i)} (\boldsymbol{\sigma}^i_j)^2 + \sum_{j=1}^{M} n_j^i (\boldsymbol{\mu}_j^i - \boldsymbol{\mu}^i)^2}{\sum_{j=1}^{M} n_j^i}.
\end{split}
\end{equation}

\section{Experiments}

\subsection{Datasets}
\label{dataset}

In the context of multi-target domain adaptation experiments, we employ GTA \cite{richter2016playing} and SYNTHIA \cite{ros2016synthia} as the synthetic dataset and the real-world datasets of Cityscapes \cite{cordts2016cityscapes}, Indian Driving Dataset \cite{varma2019idd} (IDD), ACDC \cite{sakaridis2021acdc} and DarkZurich \cite{sakaridis2020map} as the target domains. For domain adaptation methods, training involves labeled source data and unlabeled target data from different domains. We employ our merging techniques to construct a model from the trained models, employing the discussed merging methods, all without the need for direct access to this data. 

\textbf{Synthetic Datasets. }
\textit{GTA} \cite{richter2016playing} dataset comprises 24,966 synthetic images, each with a resolution of 1914×1052 pixels. These images are sourced from the video game GTA5 and come equipped with pixel-level annotations encompassing 19 categories, aligning with the annotation protocols of Cityscapes and IDD datasets.
\textit{SYNTHIA} \cite{ros2016synthia} dataset comprises 9,400 rendered images, each with a resolution of 1280×760 pixels, generated from a virtual city.

\textbf{Real-world Datasets. }
\textit{Cityscapes} \cite{cordts2016cityscapes} is a real-world dataset featuring a collection of 5,000 street scenes captured in various European cities. These scenes have been meticulously labeled, classifying objects and elements into 19 distinct categories. In our experiments, we utilize a subset of this dataset, consisting of 2,975 images for training and an additional 500 images for validation purposes.
\textit{IDD} \cite{varma2019idd} offers greater diversity compared to Cityscapes, capturing unstructured traffic scenarios on Indian roads. It boasts a total of 10,003 images, with 6,993 designated for training, 981 for validation, and 2,029 for testing.
\textit{ACDC} \cite{sakaridis2021acdc} contains 1600 training, 406 validation, and 2000 test images, evenly distributed among adverse weather conditions, including fog, night, rain and snow. \textit{DarkZurich}\cite{sakaridis2020map} contains 2416 training, 50 validation, and 151 test images specifically curated for nighttime scenarios.

\subsection{Implementation Details}
\label{impl-details}

Our proposed model merging techniques are easy to implement.
In the first phase of training single-target domain adaptation model, we leverage the state-of-the-art HRDA method \cite{hoyer2022hrda}.
We validate the effectiveness of our approach using a range of image encoders as backbones, including ResNet50 \cite{he2016deep}, ResNet101 \cite{he2016deep}, and MiT-B5 \cite{xie2021segformer}, all pre-trained on ImageNet-1K \cite{deng2009imagenet}.
Except for the backbone encoder and the factor under examination in Fig.~\ref{fig:linear-analysis}, all other training hyperparameters remain consistent with the original HRDA \cite{hoyer2022hrda} implementation.
In the second phase of model merging, we work directly with the state dictionaries of checkpoint files. For parameters, we perform a mid-point merging, and for buffers, we apply the formula presented in Eq.~\ref{buffer-merging}. 
When examining linear mode connectivity in Fig.~\ref{fig:linear-analysis}, we evenly sample values of $\lambda$ within the range of [0, 1] in Eq.~\ref{linear-merge}, using a stride of 0.1 and including both endpoints.

\begin{table}[ht]
\caption{
\textbf{Performance Comparison of Our Method and Baselines. }The mIoU (mean Intersection-over-Union) represents the average IoU across 19 categories. `Enc.' denotes the encoder architecture, with `R' representing ResNet101 and `V' indicating MiT-B5. The `Metric' column specifies whether evaluation was conducted on the Cityscapes (`C') or IDD (`I') dataset. The harmonic mean (`H'), representing adaptation ability across the two domains, is considered as the primary metric. Bold text highlights the best harmonic mean results, while underlined text indicates the second-best results. `\textdagger' signifies only merging backbones while keeping separate decode heads. 
}
    \centering
    \resizebox{0.9\linewidth}{!}{
    \begin{tblr}{
      cells = {c,},
      cell{2, 5, 8, 11, 14, 17,20,23,26,29,32, 35, 38, 41}{1,2} = {r=3}{},
      vline{2, 3, 4, 23} = {},
      cell{1}{23}={halign=l, bg=white},
      cell{2-Z}{23}={halign=l, preto=\;, bg=white},
    }
\hline[2pt, solid]
\SetRow{valign=m} Method & Enc. & \rotatebox{65}{Metric} & \rotatebox{65}{road} & \rotatebox{65}{sidewalk} & \rotatebox{65}{building} & \rotatebox{65}{wall} & \rotatebox{65}{fence} & \rotatebox{65}{pole} & \rotatebox{65}{light} & \rotatebox{65}{sign} & \rotatebox{65}{veg.} & \rotatebox{65}{terrain} & \rotatebox{65}{sky} & \rotatebox{65}{person} & \rotatebox{65}{rider} & \rotatebox{65}{car} & \rotatebox{65}{truck} & \rotatebox{65}{bus} & \rotatebox{65}{train} & \rotatebox{65}{motor} & \rotatebox{65}{bike} & \textbf{mIoU} &\\
\hline
\SetRow{rowsep=1pt} Data Comb. &R & C & 95.3 & 67.5     & 87.7     & 25.8 & 17.3  & 50.6 & 51.8  & 58.9 & 90.2 & 42.8    & 92.9 & 75.2   & 40.4  & 91.9 & 54.5  & 61.6 & 3.9   & 56.0    & 62.7 & 59.3\\
\SetRow{rowsep=2pt}& & I & 95.8 & 22.3     & 73.0       & 29.2 & 12.8  & 34.2 & 26.0    & 54.6 & 82.2 & 36.2    & 95.3 & 67.0     & 64.4  & 80.3 & 70.2  & 60.9 & 0.0     & 74.8  & 34.1 & 53.3 \\ 
\SetRow{rowsep=2pt}& & H & 95.6 & 33.5     & 79.7     & 27.4 & 14.7  & 40.8 & 34.6  & 56.6 & 86.0   & 39.2    & 94.1 & 70.9   & 49.6  & 85.7 & 61.3  & 61.2 & 0.0     & 64.0    & 44.2 & 56.2\\ 
\hline
\SetRow{rowsep=2pt} {STDA \\ (GTA $\rightarrow$ $\text{CS}$)} & R & C &  95.9 & 74.4     & 88.9     & 32.4 & 34.0    & 54.6 & 61.2  & 71.1 & 90.3 & 45.1    & 88.9 & 77.8   & 51.2  & 91.3 & 49.6  & 62.8 & 0.0     & 46.5  & 60.5 & 61.9\\
\SetRow{rowsep=2pt} & & I & 90.7 & 30.6     & 64.8     & 14.6 & 17.5  & 28.9 & 17.9  & 18.8 & 80.1 & 17.5    & 92.3 & 62.3   & 51.6  & 74.6 & 40.2  & 36.1 & 0.0     & 67.7  & 35.2 & 44.3 \\
\SetRow{rowsep=2pt} & & H &  93.3 & 43.3     & 75.0       & 20.2 & 23.1  & 37.8 & 27.7  & 29.8 & 84.9 & 25.2    & 90.6 & 69.2   & 51.4  & 82.1 & 44.4  & 45.9 & 0.0     & \textbf{55.1}  & \textbf{44.5} & 51.6\\
\hline
\SetRow{rowsep=2pt} {STDA \\ (GTA $\rightarrow$ $\text{IDD}$)} & R & C & 81.4 & 29.1     & 79.0       & 22.1 & 30.4  & 41.7 & 45.0    & 42.4 & 88.9 & 39.0      & 92.0   & 72.6   & 43.1  & 80.3 & 31.1  & 55.3 & 7.9   & 22.4  & 34.4 & 49.4 \\
\SetRow{rowsep=2pt} & & I & 96.0   & 34.7     & 75.0       & 33.3 & 22.6  & 33.6 & 24.2  & 54.0   & 84.9 & 39.8    & 95.8 & 69.0     & 66.0    & 80.2 & 69.4  & 63.7 & 0.0     & 76.0    & 28.9 & 55.1\\
\SetRow{rowsep=2pt} & & H & 88.1 & 31.6     & 76.9     & 26.6 & 25.9  & 37.2 & 31.5  & 47.5 & \underline{86.8} & 39.4    & \textbf{93.8} & 70.8   & 52.1  & 80.2 & 43.0    & \textbf{59.2} & 0.0     & 34.6  & 31.4 & 52.1\\
\hline
\SetRow{rowsep=2pt} {Ours} & R & C & 91.8 & 60.0       & 86.3     & 39.3 & 31.0    & 47.6 & 53.9  & 53.6 & 89.8 & 46.4    & 91.8 & 75.5   & 48.6  & 89.9 & 54.1  & 60.5 & 25.2  & 31.6  & 39.9 & 58.8 \\
\SetRow{rowsep=2pt} & & I & 96.4 & 40.0       & 73.4     & 34.9 & 23.3  & 31.7 & 27.9  & 53.6 & 83.7 & 44.8    & 94.5 & 67.8   & 64.1  & 77.6 & 57.6  & 51.3 & 0.0     & 72.8  & 30.0   & 54.0 \\
\SetRow{rowsep=2pt} & & H & \underline{94.0}   & \underline{48.0}       & \underline{79.3}     & \textbf{37.0}   & \underline{26.6}  & \underline{38.1} & \underline{36.8}  & \underline{53.6} & 86.6 & \underline{45.6}    & 93.2 & \underline{71.4}   & \underline{55.2}  & \underline{83.3} & \underline{55.8}  & 55.5 & 0.0     & 44.0    & 34.3 & \underline{56.3} \textbf{\textcolor{blue}{($\uparrow 4.2$)}} \\
\hline
\SetRow{rowsep=2pt} {Ours\textsuperscript{\textdagger}} & R & C &  92.0   & 59.6     & 87.9     & 39.1 & 35.7  & 50.7 & 59.6  & 60.8 & 90.1 & 45.4    & 91.3 & 77.4   & 48.8  & 88.4 & 52.6  & 59.6 & 20.3  & 33.5  & 50.6 & 60.2 \\
\SetRow{rowsep=2pt} & & I & 96.9 & 41.1     & 74.0       & 34.6 & 25.1  & 34.5 & 29.0    & 50.6 & 85.8 & 48.1    & 95.8 & 69.6   & 65.3  & 79.3 & 61.8  & 55.2 & 0.0     & 74.4  & 33.3 & 55.5\\
\SetRow{rowsep=2pt} & & H & \textbf{94.4} & \textbf{48.7}     & \textbf{80.4}     & \underline{36.7} & \textbf{29.5}  & \textbf{41.1} & \textbf{39.0}    & \textbf{55.2} & \textbf{87.9} & \textbf{46.7}    & \underline{93.5} & \textbf{73.3}   & \textbf{55.9}  & \textbf{83.6} & \textbf{56.8}  & \underline{57.3} & 0.0     & \underline{46.1}  & \underline{40.2} & \textbf{57.7} \textbf{\textcolor{blue}{($\uparrow 5.6$)}}\\
\hline
\hline
\SetRow{rowsep=2pt}Data Comb. &V & C & 88.9 & 49.7     & 90.2     & 58.8 & 46.6  & 50.9 & 60.9  & 60.3 & 89.7 & 47.4    & 88.8 & 77.7   & 45.0    & 93.5 & 77.7  & 79.4 & 69.0    & 57.6  & 65.8 & 68.3 \\
\SetRow{rowsep=2pt}& & I & 94.0   & 32.2     & 77.9     & 44.6 & 30.3  & 42.6 & 33.2  & 53.0   & 80.6 & 24.8    & 91.2 & 71.0     & 67.2  & 81.2 & 73.4  & 66.1 & 0.0     & 75.6  & 24.6 & 56.0 \\ 
\SetRow{rowsep=2pt}& & H & 91.4 & 39.1     & 83.6     & 50.7 & 36.7  & 46.4 & 43.0    & 56.4 & 84.9 & 32.6    & 90.0   & 74.2   & 53.9  & 86.9 & 75.5  & 72.1 & 0.0     & 65.3  & 35.8 & 61.5\\ 
\hline
\SetRow{rowsep=2pt}{STDA \\ (GTA $\rightarrow$ $\text{CS}$)} & V & C &  96.4 & 74.2     & 91.0       & 59.4 & 53.5  & 58.0   & 64.9  & 69.4 & 91.6 & 49.9    & 93.8 & 79.2   & 53.7  & 93.4 & 75.2  & 76.3 & 67.6  & 64.3  & 67.1 & 72.6 \\
\SetRow{rowsep=2pt}& & I & 85.8 & 10.0       & 72.1     & 30.7 & 27.4  & 34.0   & 32.9  & 53.8 & 79.9 & 36.4    & 95.2 & 65.2   & 54.2  & 80.2 & 47.2  & 48.8 & 0.0     & 72.4  & 34.9 & 50.6 \\
\SetRow{rowsep=2pt}& & H & 90.8 & 17.6     & 80.4     & 40.5 & \textbf{36.2}  & 42.9 & \underline{$43.6$}  & \underline{60.6} & 85.3 & 42.0      & \textbf{94.5} & 71.5   & \underline{54.0}    & 86.3 & 58.0    & 59.5 & 0.0     & \textbf{68.1}  & \textbf{45.9} & 59.6\\
\hline
\SetRow{rowsep=2pt}{STDA \\ (GTA $\rightarrow$ $\text{IDD}$)} & V & C & 87.2 & 30.0       & 88.9     & 53.1 & 35.3  & 52.0   & 56.6  & 49.1 & 89.7 & 47.3    & 88.4 & 74.9   & 40.0    & 91.1 & 76.7  & 64.2 & 29.5  & 15.2  & 30.5 & 57.9\\
\SetRow{rowsep=2pt}& & I & 93.0   & 4.7      & 78.0       & 42.0   & 24.8  & 44.3 & 25.9  & 59.8 & 79.7 & 24.1    & 91.0   & 62.7   & 59.6  & 78.8 & 71.8  & 75.6 & 0.0     & 63.1  & 16.7 & 52.4 \\
\SetRow{rowsep=2pt}& & H & 90.0   & 8.2      & \underline{83.1}     & \textbf{46.9} & 29.1  & \textbf{47.8} & 35.5  & 53.9 & 84.4 & 31.9    & 89.7 & 68.3   & 47.9  & 84.5 & \textbf{74.2}  & 69.4 & 0.0     & 24.5  & 21.6 & 55.0 \\
\hline
\SetRow{rowsep=2pt}{Ours} & V & C & 93.6 & 57.8     & 89.9     & 58.5 & 41.8  & 55.5 & 58.8  & 56.5 & 90.8 & 52.2    & 92.0   & 77.5   & 46.3  & 93.3 & 76.6  & 75.7 & 53.1  & 44.9  & 57.4 & 67.0 \\
\SetRow{rowsep=2pt}& & I & 93.6 & 18.9     & 76.1     & 35.2 & 29.4  & 38.7 & 32.7  & 58.1 & 82.2 & 41.2    & 93.9 & 72.5   & 63.3  & 81.0   & 63.7  & 66.6 & 0.0     & 75.2  & 34.9 & 55.6\\
\SetRow{rowsep=2pt}& & H & \textbf{93.6} & \textbf{28.5}     & 82.4     & 44.0   & 34.5  & 45.6 & 42.0    & 57.3 & \textbf{86.3} & \textbf{46.0}      & 92.9 & \textbf{74.9}   & 53.5  & \underline{86.7} & 69.6  & \underline{70.9} & 0.0     & 56.2  & \underline{43.4} & \underline{60.8} \textbf{\textcolor{blue}{($\uparrow 1.2$)}}\\
\hline
\SetRow{rowsep=2pt}{Ours\textsuperscript{\textdagger}} & V & C & 94.1 & 60.9     & 90.6     & 59.5 & 46.9  & 56.8 & 63.7  & 62.8 & 91.4 & 52.3    & 93.6 & 78.5   & 50.6  & 93.3 & 76.8  & 79.4 & 67.9  & 58.7  & 66.3 & 70.7\\
\SetRow{rowsep=2pt}& & I & 93.0   & 15.7     & 77.0       & 36.6 & 29.2  & 41.1 & 35.8  & 62.1 & 80.2 & 36.7    & 92.3 & 70.6   & 62.7  & 81.9 & 65.3  & 69.5 & 0.0     & 71.6  & 29.2 & 55.3\\
\SetRow{rowsep=2pt}& & H & \underline{93.5} & \underline{24.9}     & \textbf{83.2}     & \underline{45.3} & \underline{36.0}    & \underline{47.7} & \textbf{45.8}  & \textbf{62.4} & \underline{85.4} & \underline{43.1}    & \underline{93.0}   & \underline{74.3}   & \textbf{56.0}    & \textbf{87.2} & \underline{70.6}  & \textbf{74.1} & 0.0     & \underline{64.5}  & 40.5 & \textbf{62.1} \textbf{\textcolor{blue}{($\uparrow 2.5$)}}  \\
\hline[2pt, solid]
\end{tblr}
}
\label{table:main}
\end{table}

\subsection{Comparison with Baseline Methods}
\label{comp-baseline}

We present a comparison of our model merging-based multi-target domain adaptation approach with several baseline methods in Tab.~\ref{table:main}. 
In this experiment, we evaluate our method using GTA \cite{richter2016playing} as the source domain and two target domains, Cityscapes \cite{cordts2016cityscapes} and IDD \cite{varma2019idd}. However, our approach can easily scale to handle a greater number of target domains, should the need arise. Additionally, we assess the performance when SYNTHIA \cite{ros2016synthia} serves as the source domain, and the results for this scenario are presented in the supplementary material.

\textbf{Baseline methods} include Data Combination ("Data Comb.") approaches, where a single domain adaptation model is trained on a mixture of data from two target domains. Note that these are only presented for reference as they contradict our considerations related to data transfer bandwidth and privacy issues.
We also include Single-Target Domain Adaptation ("STDA") baselines, which involve training a single domain adaptation model for one domain and assessing its generalization to both domains. 
We evaluate our proposed methods (labeled as "Ours") which involve merging all models or merging only image backbones while maintaining separate decoding heads. 

\begin{wraptable}[22]{r}{0.4\textwidth}
\vspace{-2\baselineskip}
\caption{
    \textbf{Comparison of Our Method with State-of-the-Art Approaches.}
    Prior MTDA methods used different training methods from ours, only for reference.
    `\textdagger' signifies results reproduced by us.
}
    \centering
    \resizebox{1.00\linewidth}{!}{
    \begin{tblr}{
      cells = {c,},
      cell{1}{1} = {r=2}{},
      cell{1}{2} = {r=2}{},
      cell{1}{3} = {r=2}{},
      cell{1}{4} = {c=3}{c},
      cell{3}{1} = {r=16}{},
      cell{19}{1} = {r=2}{},
      cell{21}{1} = {r=4}{},
      cell{25}{1} = {r=2}{},
      vline{2, 3, 4} = {},
    }
    \hline[2pt, solid]
    Setting               & Method                      & Backbone & Metric &      &  \\
     &                             &          & CS     & IDD  & H. Mean    \\
    \hline
    {STDA \\ (GTA $\rightarrow$ X)}                  & BDL \cite{li2019bidirectional}                        & R101     & 41.1   & -    & -           \\
                      & AdaptSeg \cite{tsai2018learning}                    & R101     & 42.4   & -    & -           \\
                      & CLAN \cite{luo2019taking}                        & R101     & 43.2   & -    & -           \\
                      & ADVENT \cite{vu2019advent}                      & R101     & 43.8   & -    & -           \\
                      & MaxSquare \cite{chen2019domain}                   & R101     & 44.3   & -    & -           \\
                      & AdaptPatch \cite{tsai2019domain}                  & R101     & 44.9   & -    & -           \\
                      & CBST \cite{zou2018unsupervised}                        & R38      & 45.9   & -    & -           \\
                      & IntraDA \cite{pan2020unsupervised}                     & R101     & 46.3   & -    & -           \\
                      & DACS \cite{tranheden2021dacs}                        & R101     & 52.1   & -    & -           \\
                      & DAFormer \cite{hoyer2022daformer}                    & R101     & 56.0     & -    & -           \\
                      & CorDA \cite{wang2021domain}                       & R101     & 56.6   & -    & -           \\
                      & ProDA \cite{zhang2021prototypical}                       & R101     & 57.5   & -    & -           \\
                      & HRDA\textsuperscript{\textdagger} \cite{hoyer2022hrda} & R101     & 61.9   & -    & -           \\
                      & DDB\cite{chen2022deliberated}     & R101     & 62.7 & -    & -   \\
                      & DAFormer \cite{hoyer2022daformer}                    & MiT-B5   & 68.3   & -    & -           \\
                      & HRDA\textsuperscript{\textdagger} \cite{hoyer2022hrda} & MiT-B5   & 72.6   & -    & -           \\

    \hline
    DG                   & Yue et al. \cite{yue2019domain}                          & R101     & 42.1   & 42.8 & 42.4 \\
                      & Kundu et al. \cite{kundu2021generalize}                           & R101     & 53.4   & -    & -           \\
    \hline
    MTDA                  & MTDA-ITA \cite{gholami2020unsupervised}                    & R101     & 40.3   & 41.2 & 40.8 \\
                      & MT-MTDA \cite{nguyen2021unsupervised}                     & R101     & 43.2   & 44.0   & 43.6 \\
                      & CCL \cite{isobe2021multi}                         & R101     & 45.0     & 46.0   & 45.5 \\ 
                      & Coast\cite{zhang2023cooperative}  & R101     & 47.1  & 49.3 &  48.2\\
    \hline
    {MTDA
\\(Merging)} & Ours                        & R101     & 58.8   & 54.0   & \underline{56.3} \textbf{\textcolor{blue}{($\uparrow$8.1)}} \\
                      & Ours                        & MiT-B5   & 67.0     & 55.6 & \textbf{60.8} \textbf{\textcolor{blue}{($\uparrow$12.6)}} \\
    \hline[2pt, solid]
\end{tblr}
}
\label{table:sota}

\end{wraptable}
\textbf{Results} obtained using convolutional-based encoder architecture ResNet101 \cite{he2016deep} and transformer-based architecture MiT-B5 \cite{xie2021segformer} are presented in Tab.~\ref{table:main}.
Our method demonstrates a notable improvement of \textbf{\textcolor{blue}{+4.2\%}} and \textbf{\textcolor{blue}{+1.2\%}} in harmonic mean when applied to the ResNet101 \cite{he2016deep} and MiT-B5 \cite{xie2021segformer} backbones, respectively, compared to the strongest single-target domain adaptation model. 
Notably, this level of performance (56.3\% harmonic mean with ResNet101 \cite{he2016deep}) is already on par with data combination methods (56.2\% harmonic mean), and we achieve it without requiring access to any training data. 
Furthermore, we explore a more relaxed setting where only the encoder backbone is merged while decoding heads are separated for various downstream domains. This is a feasible approach as the parameters of image backbones are typically orders of magnitude larger than those of decoding heads. 
Remarkably, this configuration results in a substantial performance improvement of \textbf{\textcolor{blue}{+5.6\%}} and \textbf{\textcolor{blue}{+2.5\%}} in harmonic mean for two backbones, respectively. We also find that our merging-based method consistently achieves the best harmonic means across most categories, indicating its ability to enhance adaptation globally instead of biasing to certain categories.

\subsection{Comparison with State-of-the-Arts}
\label{comp-sota}

We begin by comparing our method with the single-target domain adaptation (STDA) on the GTA$\rightarrow$Cityscapes task, as shown in Tab.~\ref{table:sota}.
It is worthy to note that our method can be applied to any of these methods, provided they adapt to different domains using the same pretrained weights. This allows us to generalize to all target domains using a single model while keeping the relatively superior performance of STDA methods.
We also compare our methods with domain generalization approaches in Tab.~\ref{table:sota}, which aim to generalize a model trained on a source domain to multiple unseen target domains. Our approach stands out by achieving superior performance without requiring additional tricks, just by means of exploiting parameter space mode connectivity.

In the realm of multi-target domain adaptation, our method also stands out.
We eliminate the need for explicit inter-domain consistency regularization or knowledge distillation of multiple student models, enabling techniques from STDA methods like multi-resolution training to transfer to MTDA tasks. 
Therefore, we witness a significant improvement over the best published results of MTDA, while eliminating the need of access to training data.

\subsection{Extending to More Target Domains}

\begin{wraptable}[20]{R}{0.4\textwidth}
\vspace{-2\baselineskip}
    \centering
    \caption{
\textbf{Application of Our Model Merging Techniques Across Four Target Domains.} The datasets Cityscapes \cite{cordts2016cityscapes}, IDD \cite{varma2019idd}, ACDC \cite{sakaridis2021acdc}, and DarkZurich \cite{sakaridis2020map} are represented by `C', `I', `A', and `D', respectively. The mIoU of each dataset and the harmonic mean (H) is reported.}
    \resizebox{1.00\linewidth}{!}{
    \begin{tblr}{
      cells = {c,m},
      cell{1}{1} = {c=1}{c},
      cell{1}{1} = {r=2}{},
      cell{1}{2} = {c=4}{c},
      cell{1}{6} = {c=6}{c},
      vline{2, 6}={1-17}{},  
      vline{10} = {3-17}{}
    }
    \hline[2pt, solid]
    Model \# & Merging of & & & & Metric & \\[0.5ex]  
    & C & I & A & D & C  & I & A & D & H & $\Delta$ & 
    \\ \hline 1 & \ding{51} & & & & 61.3 & 47.0   & 42.1 & 16.1 & 32.4    & \textcolor{red}{-1.0\%} 
    \\ 2 & & \ding{51} & & & 51.2 & 55.4 & 37.6 & 16.2 & 31.8    & \textcolor{red}{-1.7\%}
    \\ 3& & & \ding{51} & & 44.9 & 38.5 & 42.4 & 18.6 & 31.9    & \textcolor{red}{-1.6\%}
    \\ 4 & & & & \ding{51} & 43.6 & 38.2 & 41.6 & 21.6 & 33.5    & -
    \\ \hline 
       5 & \ding{51} & \ding{51} & & & 58.8 & 54.0   & 41.9 & 17.4 & 34.2    & \textcolor{blue}{+0.8\%}
    \\ 6 & \ding{51} & & \ding{51} & & 55.1 & 45.2 & 45.2 & 20.8 & 36.2    & \textcolor{blue}{+2.8\%}
    \\ 7 & \ding{51} & & & \ding{51} & 57.5 & 45.2 & 45.8 & 24.2 & 38.9    & \textcolor{blue}{+5.5\%}
    \\ 8 & & \ding{51} & \ding{51} & & 47.1 & 49.7 & 43.6 & 21.5 & 36.1    & \textcolor{blue}{+2.7\%}
    \\ 9 & & \ding{51} & & \ding{51} & 50.0   & 49.6 & 46.2 & 25.0   & \textbf{39.3}    & \textcolor{blue}{+5.8\%}
    \\ 10 & & & \ding{51} & \ding{51} & 45.5 & 41.3 & 45.3 & 22.0   & 35.2    & \textcolor{blue}{+1.7\%}
    \\ \hline
       11 & \ding{51} & \ding{51} & \ding{51} & & 53.0   & 49.3 & 44.5 & 20.0   & 35.9    & \textcolor{blue}{+2.4\%}
    \\ 12 & \ding{51} & \ding{51} & & \ding{51} & 55.1 & 49.3 & 45.4 & 21.1 & 37.1    & \textcolor{blue}{+3.6\%}
    \\ 13 & \ding{51} & & \ding{51} & \ding{51} & 51.9 & 43.8 & 46.3 & 22.1 & 36.7    & \textcolor{blue}{+3.3\%}
    \\ 14 & & \ding{51} & \ding{51} & \ding{51} & 47.6 & 46.7 & 46.1 & 22.5 & 36.8    & \textcolor{blue}{+3.4\%} 
    \\ \hline 15 & \ding{51} & \ding{51} & \ding{51} & \ding{51} & 51.6 & 47.3 & 45.8 & 21.1 & 36.4    & \textcolor{blue}{+3.0\%}
    \\ \hline[2pt, solid]
    \end{tblr}
}
\label{table:four-domains}
\end{wraptable}

In this section, we expand the application of our model merging technique to encompass four distinct target domains: Cityscapes \cite{cordts2016cityscapes}, IDD \cite{varma2019idd}, ACDC \cite{sakaridis2021acdc}, and DarkZurich \cite{sakaridis2020map}. Each of these domains presents unique challenges and characteristics: Cityscapes \cite{cordts2016cityscapes} captures European urban settings, IDD \cite{varma2019idd} focuses on Indian road scenes, ACDC \cite{sakaridis2021acdc} is tailored to adverse weather conditions such as fog, rain, or snow, and DarkZurich \cite{sakaridis2020map} addresses night road scenes. 
We conduct a thorough evaluation of models that are trained separately for each domain as well as models created through the merging of these individually adapted models. The effectiveness of these approaches is quantified by reporting the harmonic mean of their performance across these diverse domains.
All results are presented in Tab.~\ref{table:four-domains}.

Our proposed model merging techniques demonstrate a significant improvement in performance, as illustrated in Tab.~\ref{table:four-domains}. 
While we use the method with the highest harmonic mean from separately trained models as our baseline for comparison, all the approaches based on model merging outperform it, with gains as substantial as \textcolor{blue}{+5.8\%}. 
Furthermore, despite the increasing complexity in merging models from multiple, diverse domains, we observe that the overall performance across all domains does not suffer any notable decline.

Through further analysis, we reveal that our approach is capable of simplifying domain consistency complexity. While existing methods like \cite{koh2022consistency,reddy2024towards} involve $O(n^2)$ considerations for inter-domain consistency regularization and online knowledge distillation, our approach reduces this to a more efficient $O(n)$, where $n$ represents the number of target domains considered.

As shown in supplementary material, we have also included Mapillary as a target domain and compared it previous work \cite{lee2022adas}. 

\subsection{Ablation Study}
\label{ablation}

\begin{wraptable}[14]{r}{0.4\textwidth}
\vspace{-2\baselineskip}
    \caption{\textbf{Ablation Study on Different Vision Backbones.}}
    \begin{minipage}{0.4\textwidth}
    \centering
        \begin{subtable}{1.0\linewidth}
            \centering
            \caption{ResNet101 Backbone}
            \resizebox{1.00\linewidth}{!}{
            \begin{tblr}{
              cells = {c,m},
              cell{1}{1} = {r=2}{},
              cell{1}{2} = {r=2}{},
              cell{1}{3} = {r=2}{},
              cell{1}{4} = {c=4}{c},
              cell{5}{1} = {r=2}{c},
              vline{2,4}={},
            }
            \hline[2pt, solid]
            Setting               & {Merging \\ Params.}             & {Merging \\ Buffers} & Metric  \\
             &                             &          & C     & I  & H  & $\Delta$ \\[0.45ex]
            \hline
            {STDA (G $\rightarrow$ C)} & - & - & \textbf{61.9} & 44.3 & 51.6  & \textcolor{red}{-0.5\%} \\[0.45ex]
            {STDA (G $\rightarrow$ I)} & - & - & 49.4 & \textbf{55.1} & 52.1 & - \\[0.45ex]
            \hline
            Ours & \ding{51} & \ding{55} & 58.7 & 51.3 & \textbf{54.8} & \textcolor{blue}{+2.7\%} \\ [0.45ex]
            & \ding{51} & \ding{51} & \underline{58.8} & \underline{54.0} & \textbf{56.3} & \textcolor{blue}{+4.2\%} \\
            \hline[2pt, solid]
            
        \end{tblr}
        }
        \label{subtable:resnet101}
        \end{subtable}%
    \end{minipage}
    \hfill
    \begin{minipage}{0.4\textwidth}
        \begin{subtable}{1.0\linewidth}
            \centering
            \caption{MiT-B5 Backbone}
    \resizebox{1.00\linewidth}{!}{
    \begin{tblr}{
      cells = {c,m},
      cell{1}{1} = {r=2}{},
      cell{1}{2} = {r=2}{},
      cell{1}{3} = {r=2}{},
      cell{1}{4} = {c=4}{c},
      cell{5}{1} = {r=2}{c},
      vline{2,4}={},
    }
    \hline[2pt, solid]
    Setting               & {Merging \\ Params.}             & {Merging \\ Buffers} & Metric   \\
     &                             &          & C     & I  & H & $\Delta$ \\[0.45ex]
    \hline
    {STDA (G $\rightarrow$ C)} & - & - & \textbf{72.6} & 50.6 & 59.6 & - \\[0.45ex]
    {STDA (G $\rightarrow$ I)} & - & - & 57.9 & 52.4 & 55.0 & \textcolor{red}{-4.6\%} \\[0.45ex]
    \hline
    Ours & \ding{51} & \ding{55} & 66.6 & \underline{54.9} & \underline{60.2} & \textcolor{blue}{+0.6\%} \\ [0.45ex]
    & \ding{51} & \ding{51} & \underline{67.0} & \textbf{55.6} & \textbf{60.8} & \textcolor{blue}{+1.2\%} \\
    \hline[2pt, solid]
    
\end{tblr}
}
            \label{subtable:vit}
        \end{subtable}%
    \end{minipage}
    \label{table:ablation}
\end{wraptable}

\textbf{Weight Merging and Buffer Merging.}
We conduct ablation studies on our proposed parameter merging and buffer merging methods using ResNet101 \cite{he2016deep} and MiT-B5 \cite{xie2021segformer} as the image encoders in the segmentation network \cite{hoyer2022hrda}, with results reported in Tab.~\ref{table:ablation}(a) and Tab.~\ref{table:ablation}(b), respectively.
We have observed variations in the generalization capabilities of single-target domain adaptation (STDA) models across different domains, which primarily arise from the varying diversity and quality of the target datasets used. 
Nonetheless, we select the higher harmonic means from STDA models as our baseline for comparison. 
The data in Tab.~\ref{table:ablation}(a) and Tab.~\ref{table:ablation}(b) reveal that employing a straightforward mid-point merging approach for parameters leads to an increase in generalization ability by \textcolor{blue}{+2.7\%} and \textcolor{blue}{+0.6\%}. 
Furthermore, when buffer merging is incorporated, this enhancement in performance is further amplified to \textcolor{blue}{+4.2\%} and \textcolor{blue}{+1.2\%}.
We also observe an intriguing phenomenon with the MiT-B5 backbone: the merged model outperforms the single-target adapted model when evaluating in the IDD domain. This finding implies that domain-invariant knowledge can be acquired from other domains.
These results suggest that each part of our proposed model merging technique is effective.

\textbf{Different STDA Methods \& Tasks. }
To validate the versatility of our proposed model merging method, we conducted experiments on another STDA Method ADVENT \cite{vu2019advent}. We also apply our methods on image classification tasks. Details are in the supplementary material.

\section{Conclusion}
This paper introduces a novel model merging strategy aimed at addressing the multi-target domain adaptation (MTDA) challenge without relying on training data. 
Our findings reveal that when pretrained on extensive datasets, both deep convolutional neural networks and transformer-based vision models can confine the finetuned models with the same basin in the loss landscape. 
We also emphasize the significance of buffer merging in MTDA, as buffers are key to capturing the unique features of various domains. 
The methods we propose are straightforward yet highly effective, establishing new state-of-the-art results on the MTDA benchmark. 
We anticipate that the concepts and methodologies presented in this paper will inspire future explorations in this field.

\newpage
\section*{Acknowledgements}
This research is supported by Tsinghua University – Mercedes Benz Institute for Sustainable Mobility.

\bibliographystyle{splncs04}

\newpage
\appendix

\section{Results of SYNTHIA as Source Domain}
In this section, we show more experimental results with SYNTHIA \cite{ros2016synthia} dataset as the source domain.

\textbf{Comparison with Baseline Models.} Tab.~\ref{table:supp_main} shows the performance comparison with baseline models using SYNTHIA \cite{ros2016synthia} as the source domain and two target domains, Cityscapes \cite{cordts2016cityscapes} and IDD \cite{varma2019idd}. Please note that the mIoU (mean Intersection-over-Union) is calculated across 13 categories within standard SYNTHIA \cite{ros2016synthia} evaluation protocol.

\begin{table*}[htbp!]
\centering
\caption{\textbf{Performance Comparison of Our Method and Baseline Models (SYNTHIA $\rightarrow$ X).} The mIoU* (mean Intersection-over-Union) represents the average IoU across 13 categories. `Enc.' denotes the encoder architecture, with `R' representing ResNet101 and `V' indicating MiT-B5. The `Metric' column specifies whether evaluation was conducted on the Cityscapes (`C') or IDD (`I') dataset. The harmonic mean (`H'), representing adaptation ability across the two domains, is considered as the primary metric. Bold text highlights the best harmonic mean results, while underlined text indicates the second-best results. `\textdagger' signifies only merging backbones while keeping separate decode heads.
}
    \resizebox{1.00\linewidth}{!}{
    \begin{tblr}{
      cells = {c,},
      cell{2, 5, 8, 11, 14, 17,20,23,26,29,32, 35, 38, 41}{1,2} = {r=3}{},
      row{2-4, 17-19} = {fg=gray},
      vline{2, 3, 4, 23} = {},
      cell{1}{23}={halign=l, bg=white},
      cell{2-Z}{23}={halign=l, preto=\;, bg=white},
    }
\hline[2pt, solid]
\SetRow{valign=m} Method & Enc. & \rotatebox{65}{Metric} & \rotatebox{65}{road} & \rotatebox{65}{sidewalk} & \rotatebox{65}{building} & \rotatebox{65}{light} & \rotatebox{65}{sign} & \rotatebox{65}{veg.} & \rotatebox{65}{sky} & \rotatebox{65}{person} & \rotatebox{65}{rider} & \rotatebox{65}{car}  & \rotatebox{65}{bus}  & \rotatebox{65}{motor} & \rotatebox{65}{bike} & \textbf{mIoU*} &\\
\hline
\SetRow{rowsep=1pt} Data Comb. &R & C & 87.4 & 15.2 & 86.7 & 55.0 & 57.8 & 87.2 & 92.9 & 76.9 & 44.5 & 85.5 & 35.7 & 48.3 & 63.6 & 64.4  \\
\SetRow{rowsep=2pt}& & I & 94.4 & 25.8 & 67.7 & 21.2 & 22.0 & 83.9 & 94.7 & 65.9 & 58.9 & 53.8 & 32.5 & 70.2 & 41.1 & 56.3  \\
\SetRow{rowsep=2pt}& & H & 90.8 & 19.1 & 76.0 & 30.6 & 31.9 & 85.5 & 93.8 & 71.0 & 50.7 & 66.0 & 34.0 & 57.2 & 49.9 & 60.1  \\
\hline
\SetRow{rowsep=2pt} {STDA \\ (SYNTHIA $\rightarrow$ \text{CS})} & R & C &  76.8 & 36.8 & 87.0 & 62.1 & 62.1 & 87.0 & 90.8 & 75.9 & 51.6 & 85.8 & 34.0 & 53.7 & 62.8 & 66.6  \\
\SetRow{rowsep=2pt} & & I & 57.5 & 4.1  & 54.1 & 18.8 & 35.4 & 81.3 & 94.6 & 55.8 & 49.7 & 48.6 & 23.4 & 58.7 & 20.4 & 46.3  \\
\SetRow{rowsep=2pt} & & H & 65.8 & 7.3  & 66.7 & 28.9 & \underline{45.1} & 84.0 & 92.6 & 64.3 & \underline{50.6} & 62.0 & 27.7 & \underline{56.1} & 30.7 & 54.7  \\
\hline
\SetRow{rowsep=2pt} {STDA \\ (SYNTHIA $\rightarrow$ \text{IDD})} & R & C & 84.5 & 25.3 & 83.8 & 30.6 & 48.6 & 86.4 & 92.4 & 74.6 & 32.7 & 75.8 & 31.4 & 18.8 & 32.3 & 55.2  \\
\SetRow{rowsep=2pt} & & I & 93.9 & 28.9 & 68.4 & 7.0  & 30.7 & 85.7 & 96.2 & 67.1 & 52.2 & 49.0 & 39.4 & 57.9 & 31.0 & 54.4  \\
\SetRow{rowsep=2pt} & & H & \textbf{88.9} & 27.0 & 75.3 & 11.4 & 37.7 & 86.1 & 94.3 & \underline{70.6} & 40.2 & 59.5 & \textbf{34.9} & 28.3 & 31.6 & 54.8  \\
\hline
\SetRow{rowsep=2pt} {Ours} & R & C & 85.3 & 45.5 & 86.5 & 53.4 & 60.6 & 87.7 & 92.5 & 76.9 & 42.8 & 84.4 & 30.5 & 47.4 & 53.5 & 65.1  \\
\SetRow{rowsep=2pt} & & I & 91.7 & 18.4 & 67.2 & 21.7 & 42.1 & 85.3 & 96.3 & 64.5 & 55.8 & 50.1 & 36.0 & 62.6 & 35.5 & 55.9  \\
\SetRow{rowsep=2pt} & & H & \underline{88.4} & \underline{26.2} & \underline{75.6} & \textbf{30.8} & \textbf{49.7} & \textbf{86.5} & \textbf{94.3} & 70.2 & 48.4 & \textbf{62.9} & \underline{33.0} & 53.9 & \underline{42.7} & \underline{60.2}  \textbf{\textcolor{blue}{($\uparrow$5.4)}}\\
\hline
\SetRow{rowsep=2pt} {Ours\textsuperscript{\textdagger}} & R & C &  82.7 & 43.0 & 86.3 & 58.8 & 62.2 & 87.8 & 92.5 & 77.7 & 48.9 & 84.5 & 28.3 & 52.4 & 61.5 & 66.7  \\
\SetRow{rowsep=2pt} & & I & 93.1 & 26.9 & 67.6 & 20.8 & 33.6 & 85.1 & 96.2 & 65.6 & 54.2 & 49.8 & 38.9 & 61.4 & 35.7 & 56.1  \\
\SetRow{rowsep=2pt} & & H & 87.6 & \textbf{33.1} & \textbf{75.8} & \underline{30.7} & 43.6 & \underline{86.4} & \underline{94.3} & \textbf{71.1} & \textbf{51.5} & \underline{62.6} & 32.8 &\textbf{56.5} & \textbf{45.2} & \textbf{60.9} \textbf{\textcolor{blue}{($\uparrow$6.1)}}\\
\hline
\hline
\SetRow{rowsep=2pt} Data Comb. &R & C & 85.5 & 40.0 & 88.6 & 62.6 & 58.2 & 87.4 & 89.6 & 74.1 & 31.5 & 88.0 & 55.9 & 51.9 & 62.4 & 67.3  \\
\SetRow{rowsep=2pt}& & I & 75.4 & 20.3 & 72.1 & 25.6 & 49.1 & 65.2 & 91.2 & 71.5 & 66.2 & 54.5 & 51.3 & 73.1 & 41.6 & 58.2  \\ 
\SetRow{rowsep=2pt}& & H & 80.1 & 26.9 & 79.5 & 36.4 & 53.2 & 74.7 & 90.4 & 72.8 & 42.7 & 67.3 & 53.5 & 60.7 & 49.9 & 62.5  \\
\hline
\SetRow{rowsep=2pt}{STDA \\ (SYNTHIA $\rightarrow$ \text{CS})} & V & C &  86.9 & 52.0 & 89.6 & 65.4 & 58.6 & 85.4 & 94.2 & 79.5 & 54.3 & 86.5 & 54.0 & 59.4 & 63.0 & 71.4  \\
\SetRow{rowsep=2pt}& & I & 70.0 & 5.0  & 66.2 & 28.1 & 46.8 & 85.0 & 96.2 & 58.1 & 47.3 & 57.6 & 49.4 & 66.8 & 25.5 & 54.0  \\
\SetRow{rowsep=2pt}& & H & 77.5 & 9.1  & 76.1 & \textbf{39.3} & 52.0 & 85.2 & \textbf{95.2} & 67.1 & \underline{50.5} & \textbf{69.1} & 51.6 & \underline{62.9} & 36.3 & 61.5  \\
\hline
\SetRow{rowsep=2pt}{STDA \\ (SYNTHIA $\rightarrow$ \text{IDD})} & V & C & 77.7 & 32.6 & 86.9 & 41.9 & 49.5 & 88.5 & 87.9 & 75.3 & 39.8 & 87.8 & 59.8 & 29.5 & 45.7 & 61.8  \\
\SetRow{rowsep=2pt}& & I & 89.2 & 36.7 & 71.6 & 16.4 & 56.7 & 79.9 & 90.6 & 73.9 & 67.4 & 53.6 & 64.3 & 73.8 & 39.4 & 62.6  \\
\SetRow{rowsep=2pt}& & H & 83.1 & \textbf{34.5} & 78.5 & 23.6 & 52.9 & 84.0 & 89.3 & \textbf{74.6} & 50.1 & 66.6 & \underline{62.0} & 42.1 & 42.3 & 62.2  \\
\hline
\SetRow{rowsep=2pt}{Ours} & V & C & 88.2 & 48.8 & 88.8 & 55.9 & 56.7 & 88.4 & 92.4 & 76.3 & 43.3 & 88.8 & 62.2 & 55.3 & 58.3 & 69.5  \\
\SetRow{rowsep=2pt}& & I & 88.5 & 14.1 & 72.0 & 28.0 & 56.8 & 85.0 & 94.3 & 67.0 & 58.8 & 54.6 & 61.5 & 72.0 & 36.1 & 60.7  \\
\SetRow{rowsep=2pt}& & H & \textbf{88.3} & 21.8 & \underline{79.5} & 37.3 & \underline{56.7} & \textbf{86.7} & \underline{93.3} & 71.3 & 49.8 & \underline{67.6} & 61.8 & 62.5 & \underline{44.6} & \underline{64.8} \textbf{\textcolor{blue}{($\uparrow$2.6)}}  \\
\hline
\SetRow{rowsep=2pt}{Ours\textsuperscript{\textdagger}} & V & C & 88.4 & 49.4 & 89.2 & 63.0 & 57.9 & 88.8 & 94.1 & 78.3 & 49.6 & 88.7 & 60.9 & 59.0 & 61.0 & 71.4  \\
\SetRow{rowsep=2pt}& & I &87.6 & 14.3 & 72.1 & 27.3 & 56.9 & 82.4 & 92.0 & 70.4 & 63.6 & 54.5 & 63.2 & 74.5 & 38.9 & 61.4  \\
\SetRow{rowsep=2pt}& & H & \underline{88.0} & \underline{22.2} & \textbf{79.7} & \underline{38.1} & \textbf{57.4} & \underline{85.5} & 93.0 & \underline{74.2} & \textbf{55.7} & 67.6 & \textbf{62.0} & \textbf{65.9} & \textbf{47.5} & \textbf{66.0} \textbf{\textcolor{blue}{($\uparrow$3.8)}}\\

\hline[2pt, solid]
\end{tblr}
} 

\label{table:supp_main}
\end{table*}

We can deduce similar conclusions based on the experimental results of the SYNTHIA \cite{ros2016synthia} dataset in Tab.~\ref{table:supp_main}, which show the the broad applicability of our approach. 
Our model merging method with SYNTHIA\cite{ros2016synthia} dataset as source domain demonstrates a notable improvement of \textbf{\textcolor{blue}{+5.4\%}} and \textbf{\textcolor{blue}{+2.6\%}} in harmonic mean when applied to the ResNet101 \cite{he2016deep} and MiT-B5 \cite{xie2021segformer} backbones, respectively, compared to the strongest single-target domain adaptation model.

\begin{table}[h!]
    \centering
    \caption{
        \textbf{Comparison of Our Method with State-of-the-Art Approaches (SYNTHIA $\rightarrow$ X).}
        We present the performance of various methods in different settings, including single-target domain adaptation (STDA), domain generalization (DG), multi-target domain adaptation with access to multiple target domains simultaneously (MTDA), and MTDA (merging) which stands for our proposed setting.
        `\textdagger' signifies results reproduced by us.
    }
    \resizebox{0.90\linewidth}{!}{
    \begin{tblr}{cells = {c,},
      cell{1}{1} = {r=2}{},
      cell{1}{2} = {r=2}{},
      cell{1}{3} = {r=2}{},
      cell{1}{4} = {c=3}{},
      cell{3}{1} = {r=13}{},
      cell{3}{4} = {c},
      cell{4}{4} = {c},
      cell{5}{4} = {c},
      cell{6}{4} = {c},
      cell{7}{4} = {c},
      cell{8}{4} = {c},
      cell{9}{4} = {c},
      cell{10}{4} = {c},
      cell{11}{4} = {c},
      cell{11}{5} = {c},
      cell{11}{6} = {c},
      cell{12}{4} = {c},
      cell{12}{5} = {c},
      cell{12}{6} = {c},
      cell{13}{4} = {c},
      cell{14}{4} = {c},
      cell{14}{5} = {c},
      cell{14}{6} = {c},
      cell{15}{4} = {c},
      cell{15}{5} = {c},
      cell{15}{6} = {c},
      cell{16}{1} = {r=2}{},
      cell{16}{4} = {c},
      cell{16}{5} = {c},
      cell{16}{6} = {c},
      cell{17}{4} = {c},
      cell{18}{1} = {r=3}{},
      cell{18}{4} = {c},
      cell{18}{5} = {c},
      cell{18}{6} = {c},
      cell{19}{4} = {c},
      cell{19}{5} = {c},
      cell{19}{6} = {c},
      cell{20}{4} = {c},
      cell{20}{5} = {c},
      cell{20}{6} = {c},
      cell{21}{1} = {r=2}{},
      cell{21}{4} = {c},
      cell{21}{5} = {c},
      cell{21}{6} = {c},
      cell{22}{4} = {c},
      cell{22}{5} = {c},
      cell{22}{6} = {c},
      vline{2, 3, 4} = {},
    }
    \hline[2pt, solid]
    Setting               & Method                      & Backbone & Metric &      &  \\
     &                             &          & CS     & IDD  & H. Mean    \\
    \hline
    {STDA \\ (SYNTHIA $\rightarrow$ X)}              & MaxSquare \cite{chen2019domain}                   & R101     & 45.8   & -    & -           \\
                      & AdaptSeg \cite{tsai2018learning}                    & R101     & 46.7   & -    & -           \\
                      & CLAN \cite{luo2019taking}                           & R101     & 47.8   & -    & -           \\
                      & ADVENT \cite{vu2019advent}                          & R101     & 47.8   & -    & -           \\
                      & IntraDA \cite{pan2020unsupervised}                  & R101     & 48.9   & -    & -           \\
                      & DACS \cite{tranheden2021dacs}                       & R101     & 54.8   & -    & -           \\
                      & CorDA \cite{wang2021domain}                         & R101     & 62.0   & -    & -           \\
                      & ProDA \cite{zhang2021prototypical}                  & R101     & 62.8   & -    & -           \\
                      & HRDA\textsuperscript{\textdagger} \cite{hoyer2022hrda} & R101     & 66.6   & -    & -  \\
                      & HRDA\textsuperscript{\textdagger} \cite{hoyer2022hrda} & R101     &  -     & 54.4    & -  \\
                      & DAFormer \cite{hoyer2022daformer}                      & MiT-B5   & 67.4   & -       & -       \\
                      & HRDA\textsuperscript{\textdagger} \cite{hoyer2022hrda} & MiT-B5   & 71.4   & -    & -    \\
                      & HRDA\textsuperscript{\textdagger} \cite{hoyer2022hrda} & MiT-B5   & -  & 62.6    & -    \\
    \hline
    DG                & Yue et al. \cite{yue2019domain}                          & R101     & 44.3   & 41.2 & 42.7 \\
                      & Kundu et al. \cite{kundu2021generalize}                  & R101     & 60.1   & -    & -     \\
    \hline
    MTDA              & MTDA-ITA \cite{gholami2020unsupervised}                   & R101     & 42.7   & 39.4   & 41.0 \\
                      & MT-MTDA \cite{nguyen2021unsupervised}                     & R101     & 45.2   & 42.2   & 43.6 \\
                      & CCL \cite{isobe2021multi}                                 & R101     & 48.1   & 44.0   & 46.0 \\ 
    \hline
    {MTDA\\(Merging)} & Ours                        & R101     & 65.1     & 55.9   & \underline{60.2} \textbf{\textcolor{blue}{($\uparrow$14.2)}} \\
                      & Ours                        & MiT-B5   & 69.5     & 60.7   & \textbf{64.8} \textbf{\textcolor{blue}{($\uparrow$18.8)}} \\
    \hline[2pt, solid]
\end{tblr}
}

\label{table:supp_sota}
\end{table}

Remarkably, the level of performance (60.2\% harmonic mean with ResNet101 \cite{he2016deep}) has already exceeded \textit{Data Combination} methods (60.1\% harmonic mean), and we achieve it without requiring access to any training data. 
When considering the relaxed setting where only the encoder backbone is merged while decoding heads are separated for various downstream domains, we achieve a substantial performance improvement of \textbf{\textcolor{blue}{+6.1\%}} and \textbf{\textcolor{blue}{+3.8\%}} in harmonic mean for two backbones, respectively.

\textbf{Comparison with SoTA Models.} As shown in Tab.~\ref{table:supp_sota}, our method achieves remarkable results, outperforming many approaches in both MTDA and DG. Utilizing the ResNet101 \cite{he2016deep} backbone, we observe a substantial improvement, achieving a \textbf{\textcolor{blue}{+14.2\%}} harmonic mean increase over the best published results. This improvement further rises to \textbf{\textcolor{blue}{+18.8\%}} with transformer-based backbones. The outstanding performance with SYNTHIA \cite{ros2016synthia} dataset as the source domain demonstrates the broad applicability of our merging method.

\begin{table}[htbp!]
\caption{\textbf{Comparison of Our Method with Data Combination Across Four Target Domains.} The datasets Cityscapes, IDD, ACDC, and DarkZurich are represented by `C', `I', `A', and `D'. }
\centering
\renewcommand{\arraystretch}{1.3}
\resizebox{0.8\linewidth}{!}{
\begin{tabular}{>{\centering\arraybackslash}p{2cm}|>{\centering\arraybackslash}p{2cm}|>{\centering\arraybackslash}p{1cm}|>{\centering\arraybackslash}p{1cm}|>{\centering\arraybackslash}p{1cm}|>{\centering\arraybackslash}p{1cm}|>{\centering\arraybackslash}p{1cm}}
\toprule
Setting & Encoder & C & I & A & D & H \\ \hline
Data Comb. & ResNet101 & 57.8 & 53.9 & 38.8 & 13.2 & 29.1 \\ \hline
\textbf{Ours} & ResNet101 & \multicolumn{1}{>{\centering\arraybackslash}p{1cm}|}{51.6} & \multicolumn{1}{>{\centering\arraybackslash}p{1cm}|}{47.3} & \multicolumn{1}{>{\centering\arraybackslash}p{1cm}|}{45.8} & \multicolumn{1}{>{\centering\arraybackslash}p{1cm}|}{21.1} & \multicolumn{1}{>{\centering\arraybackslash}p{2cm}}{\textbf{36.4} \textbf{\textcolor{blue}{($\uparrow$7.3)}}} \\ \bottomrule
\end{tabular}}
\label{tab:four-combine}
\end{table}

\textbf{Comparison with Data Combination Across Four Target Domains.} As shown in Tab.~\ref{tab:four-combine}, we compared our proposed method in scenarios where target domains are diverse. We used the GTA as the source dataset and selected four datasets as target domains to compare our merging method with the approach of combining the data from four domains for training. The results demonstrate that our proposed model merging techniques achieve \textbf{\textcolor{blue}{+7.3\%}} harmonic mean improvement in performance.

\begin{figure*}
    \centering
    \includegraphics[width=1.0\linewidth]{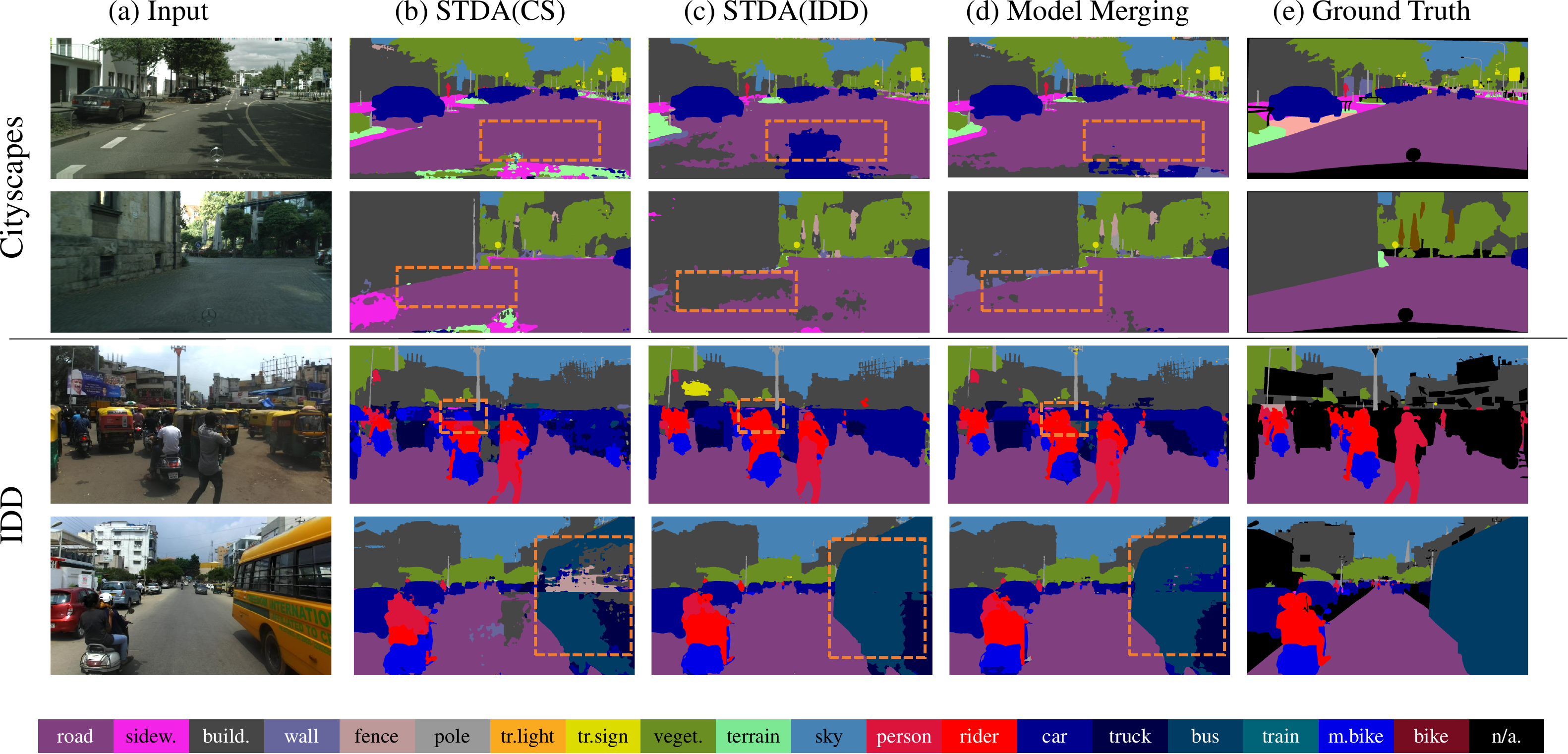}
    \caption{
    \textbf{Visualization results for GTA to Cityscapes and IDD.}
    (a) Test images from Cityscapes and IDD. We visualize results of (b) single-target domain adaptation (STDA) trained on Cityscapes target, (c) single-target domain adaptation (STDA) trained on IDD target, (d) our model merging method. (e) Ground-truth segmentation maps.}
    \label{fig:visual-compare}
\end{figure*}

\section{Visualization Results}
The qualitative comparison between different baselines and the proposed model merging method are provided in Fig.\ref{fig:visual-compare}.

As illustrated in Figure \ref{fig:visual-compare}, our proposed merging method always ensures that the combined output retains the superior predictive aspects of two separate models. 
The dotted boxes in the top two rows of this figure highlight the STDA model's proficiency in road classification within the Cityscapes dataset when Cityscapes is the target, contrasting with its less effective performance on the IDD dataset. 
However, the application of our method markedly enhances the road classification results. 
In a similar vein, the dotted boxes in the bottom two rows of Figure \ref{fig:visual-compare} showcase the STDA model's adeptness in identifying 'riders' and 'buses' in the IDD dataset when IDD is the target, as opposed to its lesser performance with Cityscapes as the target. Post-merging, the model demonstrates significantly improved classification in these categories.

\section{Results of Mapillary as Target Domain}
As shown in Table~\ref{tab:Mapillary}, we have included Mapillary as a target domain and compared it to the previous work~\cite{lee2022adas}. From the experimental results, it is clear that the domain gap between C, I, and M is smaller than that of the four selected datasets in main content (e.g., Ours (G$\rightarrow$C,I) surprisingly scores 58.6 on Mapillary despite not being adapted to it). 

\begin{table}[tp]
\centering
\caption{Model Merging Across Cityscapes(C), IDD(I) and Mapillary(M)}
\resizebox{0.50\linewidth}{!}{
\begin{tblr}{ 
  cells = {c,},
  cell{1}{1} = {r=2}{c},
  cell{1}{2} = {r=2}{c},
  cell{1}{3} = {c=4}{c},
  cell{3}{1} = {r=4}{c},
  cell{7}{1} = {r=4}{c},
  vline{2-3} = {}
}
\hline[1pt, solid]
\textbf{Method} & \textbf{Setting} & \textbf{Metric} &            &            &                  \\
                &                  & \textbf{C}      & \textbf{I} & \textbf{M} & \textbf{H. Mean} \\
\hline
ADAS \cite{lee2022adas}           & G→C,I         & 45.8            & 46.3       & -          & 46.0             \\
                & G→C,M         & 45.8            & -          & 49.2       & 47.4             \\
                & G→I,M         & -               & 46.1       & 47.6       & 46.8             \\
                & G→C,I, M      & 46.9            & 47.7       & 51.1       & 48.5             \\
\hline
Ours            & G→C,I         & 58.8            & 54.0       & 58.6       & 57.0             \\
                & G→C,M         & 60.1            & 46.5       & 59.1       & 54.5             \\
                & G→I,M         & 51.2            & 53.3       & 58.8       & 54.3             \\
                & G→C,I, M      & 58.4            & 53.2       & 59.3       & 56.8             \\
\hline[1pt, solid]

\end{tblr}
}
\label{tab:Mapillary}
\end{table}

\section{Different STDA Methods}
We verify if our proposed merging method works with other STDA methods by adopting it to another STDA Method, ADVENT \cite{vu2019advent}.
Different from teacher-student self-training frameworks, this method leverages adversarial entropy minimization to mitigate the domain gap.
According to results shown in Tab.~\ref{table:ablation_stda}, we can deduce that our model merging technique works well with it.
In fact, pre-training, which is the key to the empirical mode connectivity, has shown to pull the boundaries of NLP tasks\cite{qin2022exploring} closer, and {help in transfer learning\cite{neyshabur2020being}}.
We report that pre-training works similarly w.r.t. generic pre-trained vision backbones, and naturally could apply to another STDA methods.


\begin{table}
\centering
\caption{Verification on another STDA Method, ADVENT \cite{vu2019advent}.}
\label{table:ablation_stda}
\resizebox{0.40\linewidth}{!}{
\begin{tabular}{cccc}\\\toprule  
Setting & C & I & H  \\\midrule
ADVENT (G $\rightarrow$ C)&43.5 & 33.0 &37.5\\  \midrule
ADVENT (G $\rightarrow$ I) &37.1 & 39.8 &38.4\\  \midrule
ADVENT (Merging) &40.5 & 39.6 &\textbf{40.1}\\  \bottomrule
\end{tabular}
}
\end{table} 

\section{Extension to Image Classification Tasks}
In this section, we demonstrate the effectiveness of our model merging method applied to image classification tasks.

\begin{figure}
    \centering
    \includegraphics[width=1.0\linewidth]{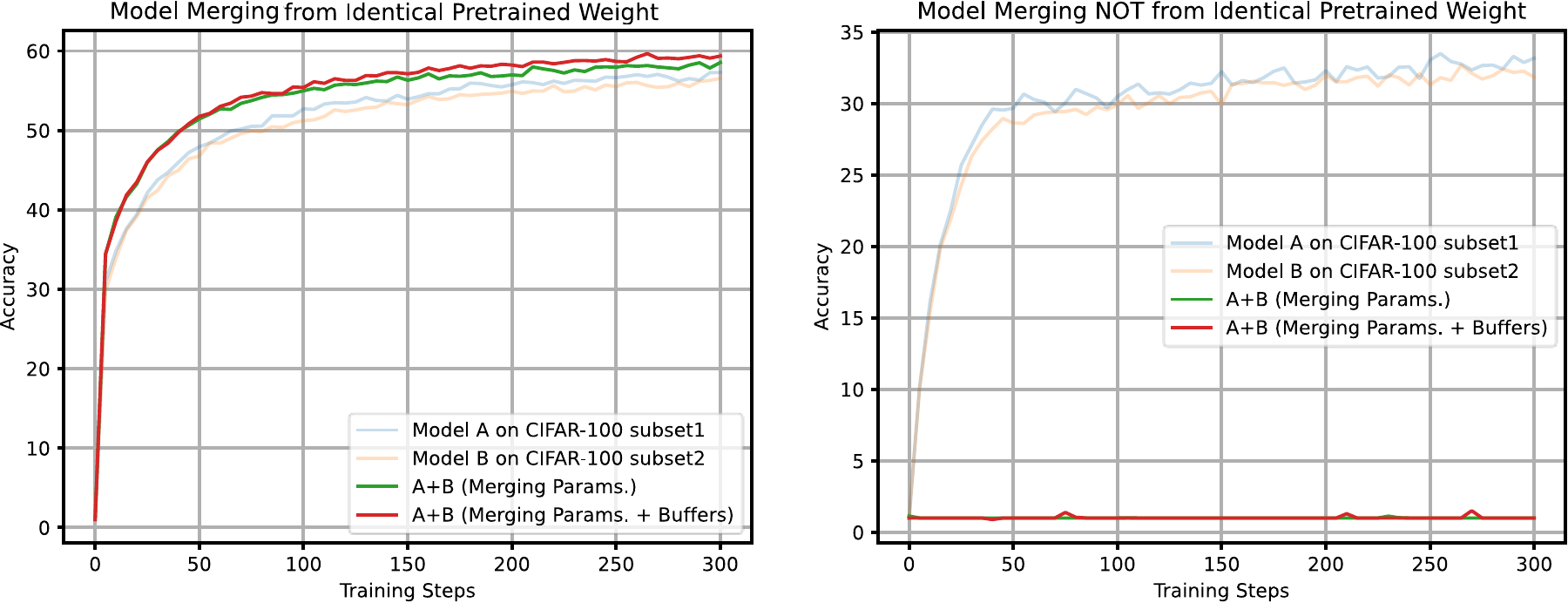}
    \caption{Model merging results on CIFAR-100 Classification.}
    \label{fig:cifar}
\end{figure}

By dividing the CIFAR-100 classification dataset into two distinct, non-overlapping subsets, we independently train two ResNet50 models, labeled A and B, on these subsets. This training was conducted either from a common set of pretrained weights or from two sets of randomly initialized weights. The performance outcomes for models A and B are illustrated in Fig.~\ref{fig:cifar}, represented by dimmed blue and yellow lines, respectively. The results indicate that models merged from a starting point of identical pretrained weights outperform those trained on any single subset. Conversely, when beginning with randomly initialized weights, individual models exhibit learning capabilities, whereas the merged model's performance is akin to making random guesses.

\begin{figure}
    \centering
    \includegraphics[width=1.0\linewidth]{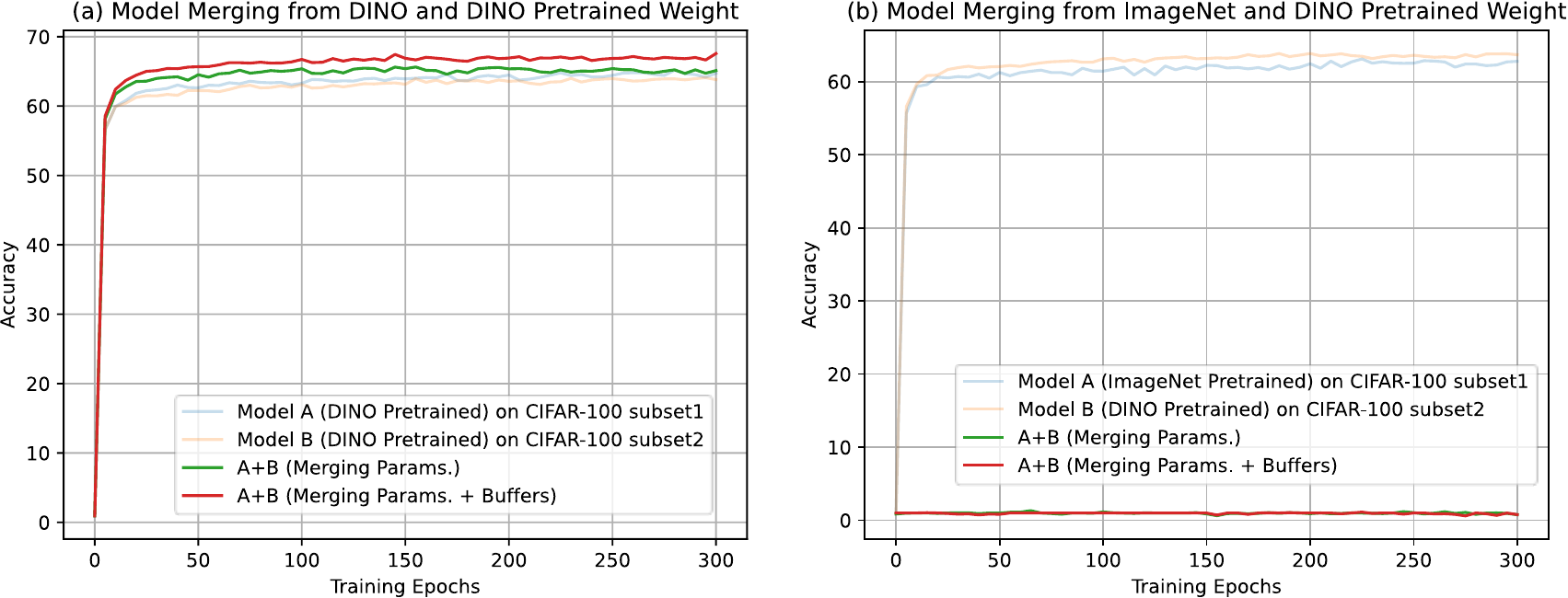}
    \caption{Results on CIFAR-100 Classification with ImageNet and DINO Pretrained Weight.}
    \label{fig:dino}
\end{figure}

Random initialization would break the linear averaging technique, while same pre-trained backbones might work. We verified this conclusion on another pre-trained weight. Results in Fig~\ref{fig:dino} indicate that DINO pretraining and ImageNet pretraining have different loss landscapes in the model's parameter space. Model merging must be conducted within the same loss landscape.

\end{document}